\documentclass{article}

\usepackage{PRIMEarxiv}
\usepackage[utf8]{inputenc}
\usepackage[T1]{fontenc}
\usepackage[colorlinks=true]{hyperref}
\hypersetup{
  linkcolor=black,
  citecolor=black,
  urlcolor=[RGB]{5,115,185}
}
\usepackage{url}
\usepackage{graphicx}
\usepackage{float}
\usepackage{amsmath,amssymb,amsfonts}
\usepackage{mathrsfs}
\usepackage{bm}
\usepackage{algpseudocode}
\usepackage{stmaryrd}
\usepackage{nicefrac}
\usepackage{microtype}
\usepackage{mathtools}
\usepackage{authblk}
\usepackage{fancyhdr}
\usepackage{booktabs}
\usepackage{enumitem}
\usepackage{multirow}
\usepackage{footmisc}
\usepackage[skip=6pt]{caption}
\usepackage[table,dvipsnames]{xcolor}
\usepackage{pgfplots}
\pgfplotsset{compat=1.16}
\usepackage{pgfplotstable}
\usepackage{xcolor, colortbl}
\usepackage{algorithm}
\usepackage{algpseudocode}  
\usepackage{titlesec}
\usepackage{amsthm}
\usepackage{tabularx}
\usepackage{etoolbox}
\usepackage{verbatim}
\usepackage{xparse}
\usepackage{listings}
\usepackage{xcolor}
\usepackage{tikz}
\usetikzlibrary{calc}
\usepackage{tabularx}

\usepackage{subcaption}
        
\usepackage{xltabular}
\usepackage{booktabs}
\usepackage{multirow}

\definecolor{TextColor}{RGB}{54,95,145}
\definecolor{DarkGrey}{RGB}{169,169,169}
\definecolor{LightGray}{rgb}{0.8,0.8,0.8}
\definecolor{LightCyan}{rgb}{0.74,0.83,0.9}
\definecolor{DarkBlue}{rgb}{0,0.28,0.67}
\definecolor{blizzardblue}{rgb}{0.67, 0.9, 0.93}
\definecolor{inchworm}{rgb}{0.7, 0.93, 0.36}
\definecolor{coralred}{rgb}{1.0, 0.25, 0.25}
\definecolor{celadon}{rgb}{0.67, 0.88, 0.69}

\titlespacing{\section}{0pt}{8pt}{4pt}
\titlespacing{\subsection}{0pt}{6pt}{3pt}
\titlespacing{\subsubsection}{0pt}{4pt}{2pt}
\titlespacing{\paragraph}{0pt}{2pt}{1pt}
\setlist[itemize]{topsep=4pt, partopsep=0pt, itemsep=2pt, parsep=0pt}
\setlist[enumerate]{topsep=4pt, partopsep=0pt, itemsep=2pt, parsep=0pt}

\newcommand{\fixedimage}[1]{\fixedimage[width=10cm, height=7cm, keepaspectratio]{#1}}

\newtheoremstyle{tightremark}
  {4pt}    
  {4pt}    
  {\itshape} 
  {}       
  {\bfseries} 
  {.}      
  {.5em}   
  {}       

\theoremstyle{tightremark}

\newcolumntype{L}{>{\raggedright\arraybackslash}X}   
\newcolumntype{Y}{>{\centering\arraybackslash}X}

\newcolumntype{Z}{>{\raggedright\arraybackslash}p{3.5cm}}
\newcolumntype{A}[1]{>{\hspace*{-#1}\centering\arraybackslash}X}
\newcolumntype{B}{>{\centering\arraybackslash}p{5cm}}
\renewcommand{\arraystretch}{1.3}  

\numberwithin{equation}{section}

\usepackage{tikz}

\DisableLigatures{encoding = *, family = *}

\newcommand{\sherpa}{\href{https://www.sherpa.ai/}{\textcolor{DarkBlue}{Sherpa.ai} }}

\titleformat{\subsection}
  {\normalfont\normalsize\bfseries}{\thesubsection}{1em}{}

\titleformat{\subsubsection}
  {\normalfont\normalsize\bfseries\itshape}{\thesubsubsection}{1em}{}

\makeatletter
\patchcmd{\@begintheorem}{\textit}{\textbf}{}{}
\makeatother

\title{\textcolor{black}{Towards the Next Frontier of LLMs, Training on Private Data: A Cross-Domain Benchmark for Federated Fine-Tuning}}

\author{%
  {\LARGE \href{https://sherpa.ai/}{Sherpa.ai}}\\
  research@sherpa.ai
}

\fancypagestyle{firstpagestyle}{%
  \fancyhf{}%
  \fancyhead[R]{\includegraphics[scale=1.0]{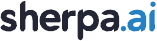}} 
  \fancyfoot[C]{\thepage}%
}

\begin{document}
\maketitle
\thispagestyle{firstpagestyle}

\begin{abstract}

The recent success of large language models (LLMs) has been largely driven by vast public datasets. However, the next frontier for LLM development lies beyond public data. Much of the world’s most valuable information is private, especially in highly regulated sectors such as healthcare and finance, where data include patient histories or customer communications. Unlocking this data could represent a major leap forward, enabling LLMs with deeper domain expertise and stronger real-world utility. Yet, these data cannot be shared because they are distributed across institutions and constrained by privacy, regulatory, and organizational barriers. Moreover, institutional datasets are typically non-independent and identically distributed (non-IID), differing across sites in population characteristics, data modalities, documentation patterns, and task-specific label distributions.

In this paper, we demonstrate a practical approach to unlocking private and distributed institutional data for LLM adaptation through federated collaboration across data silos. Built on the \sherpa Federated Learning platform, our framework enables nodes to jointly fine-tune a shared LLM without exchanging private data. We evaluate this approach through a cross-domain benchmark in healthcare and finance, using four closed-ended question answering and classification datasets: MedQA, MedMCQA, FPB, and FiQA-SA. We compare three parameter-efficient fine-tuning (PEFT) strategies—LoRA, QLoRA, and IA3—across pretrained backbones under non-IID settings reflecting institutional data heterogeneity. Our results show that federated fine-tuning performs close to centralized training and outperforms isolated single-institution learning. From a Green AI perspective, QLoRA and IA3 improve efficiency with limited accuracy degradation, supporting federated PEFT as a viable approach for adapting LLMs where data cannot be shared.

\end{abstract}


\begin{figure}[ht]
    \centering
    \includegraphics[width=0.64\linewidth]{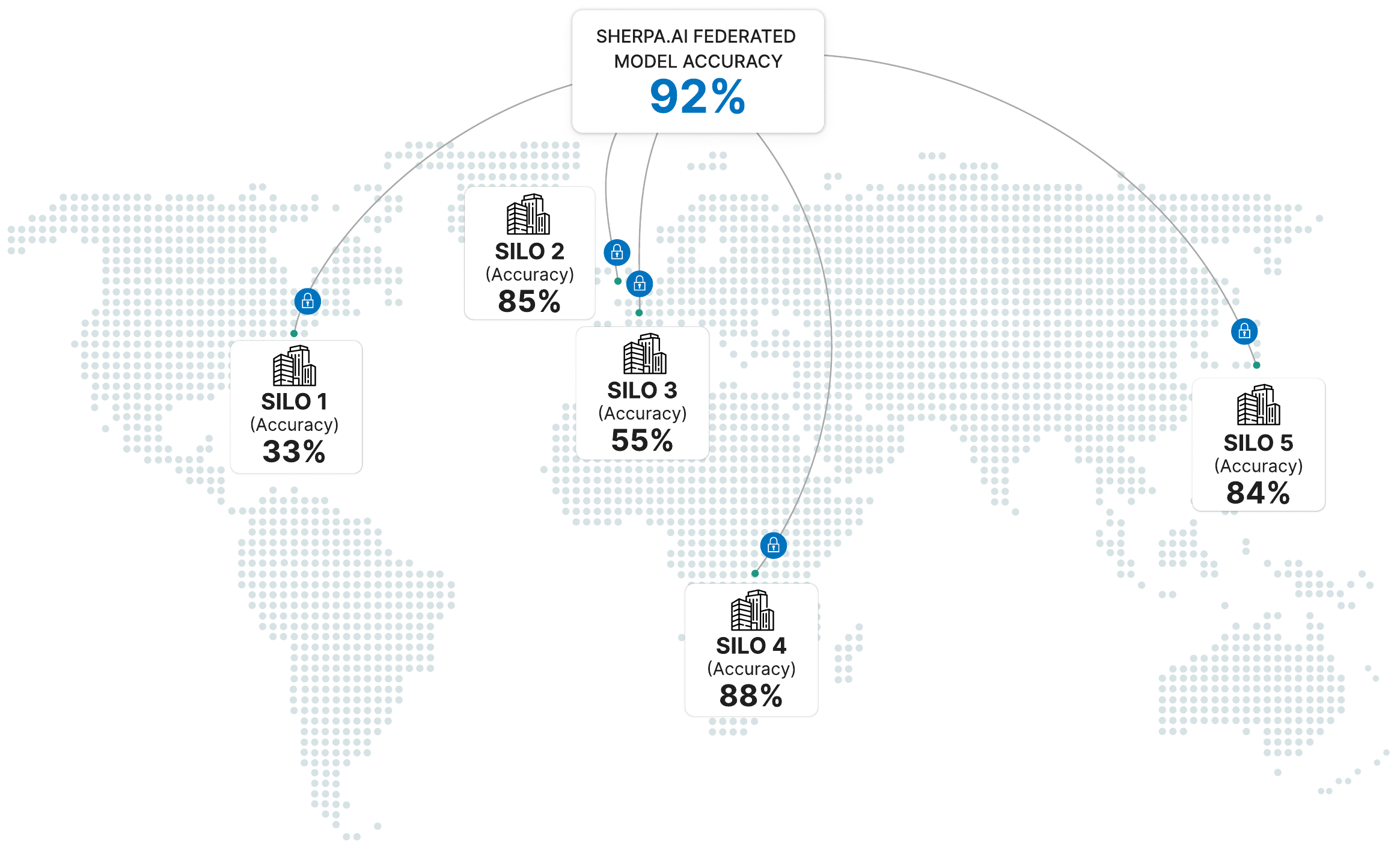}
    \caption{Global map illustrating institution-level FiQA-SA accuracies for the best federated model.}
    \label{fig:world_accuracy_comparison}
\end{figure}

\section{Introduction}
\label{sec:introduction}

Large language models (LLMs) have become a strong foundation for a wide range of natural language processing tasks, including classification, closed-ended question answering (QA), instruction following, and domain-specific reasoning in areas such as medicine and finance~\cite{ouyang2022training,liu2024large,lin2025open}. In practice, however, high performance in specialized domains still depends on task-specific adaptation through fine-tuning, i.e., further training a pretrained model on labeled task data so that it better captures the terminology and reasoning patterns of the target domain. For example, a general-purpose LLM may require additional fine-tuning to answer medical exam questions correctly or to classify the sentiment of financial texts. Parameter-efficient fine-tuning (PEFT) methods such as Low-Rank Adaptation (LoRA)~\cite{hu2022lora}, Quantized Low-Rank Adaptation (QLoRA)~\cite{dettmers2023qlora}, and infused adapters by inhibiting and amplifying inner activations (IA3)~\cite{liu2022few} have rendered model adaptation substantially practical by reducing the number of trainable parameters and memory requirements while retaining competitive performance. Thus, PEFT has become a natural choice for adapting LLMs in real-world environments with limited computational resources.

This is especially relevant in high-stakes domains such as medicine and finance, where specialized terminology, task formats, and reasoning patterns differ substantially from general-purpose pretraining corpora. Recent work (e.g., ~\cite{liu2024large,rao2025llms}) has shown that centralized fine-tuning can yield strong performance in both domains, as evidenced by large-scale medical and financial evaluation efforts. However, the centralized setting assumes that all task data can be pooled in a single location, which is often unrealistic in practice. In many real deployments, data are distributed across institutions (nodes) and cannot be shared freely due to privacy constraints, confidentiality requirements, or regulatory restrictions.

Federated learning (FL)~\cite{mcmahan2017communication} provides a natural alternative by allowing multiple nodes to collaboratively fine-tune a shared model without exchanging raw data (see Figure~\ref{fig:world_accuracy_comparison}). This paradigm is particularly attractive for LLM adaptation in sensitive domains, but it also introduces new challenges, including non-independent and identically distributed (non-IID) data across nodes, which can degrade model performance, as well as communication overhead and memory constraints. Although recent studies such as FedLLM-Bench~\cite{ye2024fedllm} and FlowerTune~\cite{gao2025flowertune} have advanced the benchmarking of federated LLM fine-tuning, the literature still lacks sufficiently controlled comparisons across training scenarios, domains, task families, and efficiency dimensions under the same experimental protocol.

\subsection{Motivation}

Although recent work has demonstrated the potential of centralized~\cite{wang2025finlora,lobo2025impact,christophe2024med42,fatemi2024comparative,wang2024finlora} and federated LLM fine-tuning~\cite{gao2025flowertune,ye2024fedllm,yan2025federated,chen2025flow}, their comparative behavior across domains, benchmarks, and PEFT strategies remains incompletely understood. Existing studies often focus on a single domain, a single benchmark, or a single adaptation strategy, making it difficult to determine how different PEFT methods behave across training paradigms under comparable conditions.

This lack of controlled comparison is particularly important in sensitive domains such as medicine and finance, where model deployment must be assessed not only in terms of predictive quality, but also with respect to practical efficiency. In these environments, communication cost, memory footprint, and robustness to non-IID data can be as relevant as accuracy itself when deciding whether a fine-tuning strategy is viable in practice.

These considerations motivate the benchmark presented in this paper. We aim to provide a controlled comparison of centralized, single-institution, and federated PEFT across multiple domains and task types, while also incorporating efficiency metrics that are critical for realistic deployment.

\subsection{Contribution}

This paper does not propose a new FL algorithm or optimization method. Instead, it presents an experimental benchmark of PEFT for LLMs under centralized, single-institution, and federated settings. The main contributions of this benchmark are as follows:

\begin{itemize}
    \item We compare PEFT-based LLM adaptation across three training scenarios: single-institution, centralized, and federated learning. All scenarios are evaluated under a common experimental protocol using three representative PEFT methods: LoRA, QLoRA, and IA3 adapters.
    
    \item We evaluate the different methods and scenarios across two task families, namely closed-ended QA and classification, and two sensitive domains, medicine and finance, enabling a controlled cross-domain and cross-task comparison.
    
    \item We complement accuracy results with efficiency metrics, namely communication cost and memory footprint, and explicitly simulate non-IID federated conditions using the Dirichlet~\cite{jimenez2024fedartml} partition protocol to create the nodes.
\end{itemize}

The remainder of this paper is organized as follows: Section~\ref{sec:problem-form} presents the problem formulation. Section~\ref{sec:ml-solution} describes the privacy-preserving machine learning (ML) solution, including the LLM fine-tuning setting and the fundamentals of FL. Section~\ref{sec:centralized-data} details the datasets, preprocessing pipeline, and centralized architecture, while Section~\ref{sec:fl-solution} describes the federated benchmark setup and the creation of nodes. Section~\ref{sec:experiments} reports the experimental setup and results. Section~\ref{sec:discussion} presents a discussion of the obtained results. Finally, Section~\ref{sec:conclusions} concludes the paper.

\section{Problem Formulation} \label{sec:problem-form}

This section presents the problem addressed in this paper: fine-tuning LLMs for closed-ended question answering and classification. We first introduce LLMs and LLM fine-tuning, and then formalize the supervised learning problem.


\subsection{LLMs and LLM Fine-Tuning}
\label{sec:llm}

An LLM is a neural network designed to process, generate, and reason over natural language at scale. Modern LLMs are typically based on the transformer architecture, which enables them to capture long-range dependencies and complex patterns in text through self-attention mechanisms. By pretraining on massive text corpora, these models acquire broad linguistic knowledge and general-purpose capabilities that can later be adapted to specific application domains and tasks.

\begin{figure}[ht]
	\centering
        \includegraphics[width=1.0\textwidth]{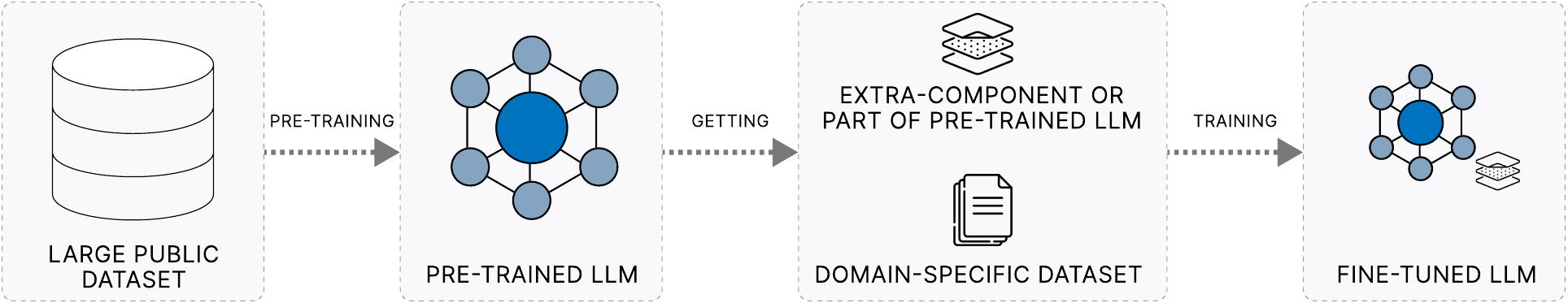}
	\caption{Overview of the simplified LLM fine-tuning process from pre-training to domain-specific adaptation.}\label{fig:finetuning_processs}
\end{figure}

LLM fine-tuning is the process of adapting a pretrained language model to a specific task 
by further optimizing it on labeled task data. As illustrated in Figure~\ref{fig:finetuning_processs}, this process starts from a pretrained LLM obtained through large-scale pretraining on a massive general-purpose corpus. The model is then adapted using a domain-specific labeled dataset, while retaining the general knowledge acquired during pretraining. During this stage, task-relevant parameters are optimized so that the model better captures the terminology, structure, and decision patterns required by the target domain. The result is a fine-tuned LLM specialized for the task.  
In this work, the target tasks are closed-ended question answering and classification, where the model must predict a correct answer option or class label from a finite set of valid outputs.

\vspace{3mm}


Although pretrained LLMs capture broad linguistic knowledge, they do not necessarily perform optimally in specialized domains such as medicine and finance. Fine-tuning addresses this limitation by exposing the model to task-specific examples, thereby improving its ability to follow the target format and produce valid domain-specific outputs.

Unlike classical ML methods based on fixed-dimensional feature vectors, LLM fine-tuning operates directly on tokenized text sequences. The model receives a prompt, and it is trained to generate the correct completion, which in our setting corresponds to a label or answer option.

Because fully fine-tuning all model parameters is often computationally expensive, this work focuses on PEFT methods, such as LoRA, QLoRA, and IA3 adapters, which update only a small set of task-specific parameters while keeping the pretrained backbone fixed or mostly frozen. These methods were selected because they are among the most widely adopted and representative approaches for efficient LLM adaptation in the recent literature, offering a practical trade-off between performance and computational cost~\cite{han2024parameter,dettmers2023qlora,hu2022lora}. This makes them particularly suitable for both centralized and federated training settings.

\subsection{Related Work}
\label{sec:related-work}

This subsection reviews the lines of work most relevant to this paper, namely, centralized and federated fine-tuning of LLMs. We emphasize prior studies on parameter-efficient adaptation, sensitive-domain applications, and realistic evaluation settings. This context helps situate the benchmark and comparative analysis presented in our work.


\subsubsection{Centralized Fine-Tuning of LLMs}
Recent work has shown that centralized fine-tuning remains a strong baseline for adapting LLMs to domain-specific tasks, especially in high-risk areas such as medicine and finance. In the medical domain, ~\cite{liu2024large} highlights the breadth of clinical evaluation settings and shows that robust assessment should go beyond a single benchmark or prompting strategy. Similarly, ~\cite{christophe2024med42} studies fine-tuning strategies for medical LLMs and compares full-parameter and parameter-efficient adaptation, showing that PEFT can be competitive while substantially reducing training cost. More recently, ~\cite{yu2025finemedlm} further demonstrates that targeted fine-tuning can improve medical reasoning ability, reinforcing the importance of specialized adaptation pipelines for healthcare tasks.

In finance, recent studies confirm that generic LLMs benefit substantially from domain adaptation. ~\cite{fatemi2024comparative} studies instruction fine-tuning for financial classification tasks and shows that fine-tuned models can outperform untuned baselines in specialized financial language understanding. ~\cite{wang2024finlora} shows that parameter-efficient and quantized adaptation is particularly attractive in finance, where local deployment and memory efficiency are often required. At a broader scale, ~\cite{rao2025llms} demonstrates that fine-tuning can yield substantial gains across a wide range of financial tasks when evaluated on a unified benchmark, while ~\cite{wang2025finlora} emphasizes the need to compare PEFT methods systematically in high-stakes financial scenarios. More generally, ~\cite{lobo2025impact} suggests that adaptation can improve task performance while also altering reasoning behavior, making careful benchmark design especially important.

Despite these advances, most centralized studies focus on a single domain, a single training regime, or a narrow set of PEFT strategies. In contrast, our work evaluates multiple PEFT methods under a unified testbed across both financial and medical closed-ended benchmarks, comparing several backbone models under the same evaluation metrics. This provides a controlled view of how centralized fine-tuning behaves across domains and creates a stronger baseline for comparison with its privacy-preserving counterpart.

\subsubsection{Federated Fine-Tuning of LLMs}
Federated fine-tuning has recently emerged as a promising alternative when domain data are distributed and cannot be pooled centrally. FedLLM-Bench~\cite{ye2024fedllm} is one of the first realistic benchmarks for FL of LLMs, showing that prior evaluations often relied on overly artificial settings and that realistic client diversity is essential for fair comparison. Independently, FlowerTune~\cite{gao2025flowertune} proposes a cross-domain benchmark for federated LLM fine-tuning spanning across general NLP, finance, medicine, and coding, hence highlighting the need for evaluation suites that go beyond a single application domain.

A complementary line of work studies algorithmic and system aspects of federated LLM adaptation. ~\cite{yan2025federated} organizes the area into parameter-sharing, knowledge-distillation, and split-learning based approaches, and clarifies the efficiency--privacy trade-offs among these families. In healthcare, ~\cite{chen2025flow} underlines the importance of domain-specific federated evaluation under non-IID client distributions, particularly in non-IID medical settings. Earlier work, such as ~\cite{hilmkil2021scaling}, established the basic feasibility of federated fine-tuning for language models. It focused on reducing memory, communication, and edge-computing costs.

Our work builds on these recent benchmarks, but addresses a different experimental question.FedLLM-Bench studies federated LLM training under instruction-tuning and preference-alignment settings, with a particular emphasis on large-scale client diversity. FlowerTune provides leaderboard-style evaluation of federated fine-tuning across various fields, evaluating multiple pretrained backbones and FL configurations. However, because FlowerTune is designed as a broad, model-agnostic benchmark, its results on demanding medical and financial tasks remain relatively far from those of the strongest proprietary or domain-specialized reference systems reported in the literature, such as GPT-4, Med-PaLM 2~\cite{medicalllmleaderboard2024}, and medical-specialized LLMs. Therefore, an additional question remains open: under a more focused sensitive-domain setup, can federated PEFT remain close to centralized fine-tuning while improving over isolated single-institution training and avoiding raw data centralization?

This paper addresses that question through a more controlled benchmark. We focus on closed-ended medical QA and financial classification, using MedQA, MedMCQA, FPB, and FiQA-SA as both task-specific adaptation and evaluation targets. We compare centralized, single-institution, and federated training under the same preprocessing pipeline, training budget, backbone models, PEFT methods, and evaluation metrics. This design allows us to quantify both the loss with respect to full centralization and the gain over isolated institutional training, while also evaluating three representative PEFT strategies: LoRA, QLoRA, and IA3 adapters. In addition to predictive performance, we report communication cost and memory footprint, enabling a joint comparison of accuracy and system efficiency on a concrete FL platform. In this sense, our contribution is complementary to previous benchmarks: it does not propose a broader leaderboard or a new FL algorithm, but provides a focused experimental benchmark for high-performing federated PEFT in sensitive domains.

\subsection{Problem Definition}
\label{sec:prob-description}

We now formalize the supervised learning setting considered in this work. Each training instance consists of a tokenized textual input and an associated target output. The input may correspond to a question or an instruction, while the target corresponds to the correct class label or answer option from a finite set of valid outputs. Since LLMs operate on token sequences rather than fixed-dimensional feature vectors, the problem is naturally expressed at the level of input tokens and discrete output labels.


Specifically, let $N$ be the total number of training instances. For each instance $i \in \{1, \dots, N\}$, we define an input sequence
\[
\mathbf{x}_i = [x_{i,1}, x_{i,2}, \dots, x_{i,p_i}],
\]
where $x_{i,j}$ denotes the $j$-th token of the input and $p_i$ is the length of the sequence for instance $i$. Depending on the task, $\mathbf{x}_i$ may represent a question, an instruction, a document-question pair, or any structured prompt provided to the language model.

The corresponding target output for each instance is denoted by $y_i$. Since we focus on closed-ended question answering and classification, $y_i$ belongs to a finite set of valid responses:
\[
y_i \in \mathcal{Y} = \{c_1, c_2, \dots, c_K\},
\]
where $\mathcal{Y}$ is the set of candidate classes or answer options, and $K = |\mathcal{Y}|$ is the number of possible outputs.

The complete dataset can be represented as
\[
\mathbf{X} = \{\mathbf{x}_1, \mathbf{x}_2, \dots, \mathbf{x}_N\}, 
\qquad
\mathbf{y} = [y_1, y_2, \dots, y_N]^\top \in \mathcal{Y}^N.
\]

Let $f_{\boldsymbol{\theta}}$ denote a pretrained large language model parameterized by $\boldsymbol{\theta}$. In PEFT, the pretrained parameters remain fully or mostly frozen, and only a small set of task-specific parameters $\boldsymbol{\phi}$ is optimized. These trainable parameters may correspond to low-rank adaptation matrices (LoRA/QLoRA) or adapter module parameters. The resulting model is denoted by
\[
f_{\boldsymbol{\theta},\boldsymbol{\phi}}.
\]

For each input $\mathbf{x}_i$, the model defines a conditional distribution over the candidate output space:
\[
P(y \mid \mathbf{x}_i; \boldsymbol{\theta}, \boldsymbol{\phi}), \qquad y \in \mathcal{Y}.
\]
The prediction for instance $i$ is then given by
\[
\hat{y}_i = \arg\max_{y \in \mathcal{Y}} P(y \mid \mathbf{x}_i; \boldsymbol{\theta}, \boldsymbol{\phi}).
\]

The objective is to learn the task-specific parameters $\boldsymbol{\phi}$ such that the model correctly maps each input sequence $\mathbf{x}_i$ to its target label or answer $y_i$. This yields a supervised learning problem over discrete outputs:
\[
\min_{\boldsymbol{\phi}} \; \mathcal{L}(\boldsymbol{\phi})
= - \sum_{i=1}^{N} \log P(y_i \mid \mathbf{x}_i; \boldsymbol{\theta}, \boldsymbol{\phi}),
\]
where $\mathcal{L}$ is the standard cross-entropy loss over the closed set of candidate answers.

Therefore, the task consists of adapting a pretrained LLM to predict the correct class or answer option from a finite output space, given an input prompt $\mathbf{x}_i$, using a PEFT strategy.

\section{Privacy-preserving ML Solution} \label{sec:ml-solution}

This section introduces the centralized and federated learning settings used in the benchmark. We adopt the notation and supervised probabilistic formulation introduced in Section~\ref{sec:prob-description}.

\subsection{Classical ML Approach}

 In the centralized setting, all training examples are pooled into a single dataset and used to adapt a pretrained large language model to the task by optimizing only a small set of task-specific parameters through PEFT.
More specifically, the available labeled data are partitioned into disjoint training, validation, and test subsets, typically after preprocessing:
\[
\{1,\ldots,N\} = I_{\mathrm{train}} \sqcup I_{\mathrm{val}} \sqcup I_{\mathrm{test}},
\]
with corresponding subsets
\[
D_{\mathrm{train}} = \{(x_i,y_i)\}_{i \in I_{\mathrm{train}}}, \qquad
D_{\mathrm{val}} = \{(x_i,y_i)\}_{i \in I_{\mathrm{val}}}, \qquad
D_{\mathrm{test}} = \{(x_i,y_i)\}_{i \in I_{\mathrm{test}}}.
\]

Let \(f_{\theta,\phi}\) denote the pretrained large language model adapted to the target task, where \(\theta\) represents the pretrained backbone parameters and \(\phi\) the trainable task-specific parameters introduced by the PEFT strategy. In the PEFT setting considered here, \(\theta\) remains fixed or mostly frozen, while only \(\phi\) is optimized.
Then, training consists of minimizing the empirical risk on \(D_{\mathrm{train}}\):
\[
J(\phi) = \frac{1}{|I_{\mathrm{train}}|}\sum_{i \in I_{\mathrm{train}}}
L\!\left(P(\cdot \mid x_i;\theta,\phi), y_i\right),
\]
where \(L\) is typically the cross-entropy loss over the finite output space. The resulting adapted model is selected using \(D_{\mathrm{val}}\) and evaluated on \(D_{\mathrm{test}}\).

\subsection{Introduction to FL}
\label{sec:fedlearning-overview}

FL enables multiple \emph{nodes} to collaboratively adapt a shared model without exchanging raw data~\cite{mcmahan2017communication}. In our setting, the shared model is a pretrained large language model with fixed base parameters \(\theta\) and trainable task-specific parameters \(\phi\). Each node \(k\) keeps its local dataset and performs local fine-tuning of the model \(f_{\theta,\phi_k}\) using its own data only. Periodically, each node transmits the trainable parameters \(\phi_k\), or updates derived from them, to a central server (the \emph{aggregator}). The server combines these local updates into a new global set of task-specific parameters \(\phi\), which is then redistributed to all nodes. This process is repeated until convergence.

FL supports data locality by ensuring that input sequences and labels remain within each institution. However, this does not by itself provide formal privacy guarantees against information leakage from model updates. Additionally, it introduces several challenges. One of the most important is the presence of non-IID data across nodes, which may cause local updates to diverge and degrade the performance of the global model. 
Thus, in this work, the focus is on non-IID scenarios, as samples and labels are distributed across different nodes according to a Dirichlet partition protocol~\cite{jimenez2024non,li2022federated,acar2021federated} (see Section~\ref{sec:nodes_creation}). In the LLM setting, the aggregation step acts on the task-specific trainable parameters \(\phi_k\), while the pretrained backbone parameters \(\theta\) remain fixed or mostly frozen.

\subsubsection{FL Paradigms}
\label{sec:fedlearning-paradigms}

In general, there are two FL paradigms depending on the different data-distribution scenarios:
\begin{itemize}
  \item \textbf{Horizontal FL (HFL):} In this paradigm, all nodes share the same input and output spaces, but each node contains different samples. In our setting, this means that all institutions fine-tune the same pretrained LLM architecture on the same task, using local datasets of the form \((\mathbf{x}_i,y_i)\), where \(\mathbf{x}_i\) is a tokenized input sequence and \(y_i \in \mathcal{Y}\) is a class or closed-ended answer option. \emph{Example}: multiple financial institutions use the same prompt structure and target label space, but each institution owns different clients, documents, or question-answer pairs.
  \item \textbf{Vertical FL (VFL):} Nodes hold complementary information about the same underlying instances. In the LLM setting, this may correspond to different institutions holding different textual or structured attributes associated with the same sample or client. \emph{Example}: multiple financial institutions collaborate on the same anti-money-laundering investigation, with one of them holding the target labels, while each institution holds different textual information associated with the same customer or case, such as transaction-related narratives, customer communications, internal compliance notes, or suspicious-activity reports.
\end{itemize}

In this work, we focus exclusively on HFL. After training is complete in an HFL setup, the resulting global adapted model is typically shared with all participating parties. This allows each node to download the final global parameters \(\phi\) and perform inference locally and independently with the model \(f_{\theta,\phi}\), without requiring further interaction or exchange of raw data.

\subsubsection{HFL}
\label{sec:fedlearning-hfl}

Under HFL, each node \(k\) has a local dataset
\[
\mathcal{D}_k
=
\{(\mathbf{x}_i^k,\,y_i^k)\}_{i=1}^{N_k},
\]
where each input sequence is of the form
\[
\mathbf{x}_i^k = [x_{i,1}^k, x_{i,2}^k, \dots, x_{i,p_i^k}^k] \in \mathcal{V}^{p_i^k},
\]
and each target satisfies
\[
y_i^k \in \mathcal{Y}.
\]
All nodes share the same vocabulary space \(\mathcal{V}\), the same output space \(\mathcal{Y}\), and the same pretrained model architecture, although the number of local samples \(N_k\) and the empirical data distribution may differ between nodes.

Training proceeds as follows:
\begin{enumerate}
  \item \textbf{Local update:} Each node \(k\) receives the current global task-specific parameters \(\phi\), initializes its local model as \(f_{\theta,\phi}\), and performs local fine-tuning on \(\mathcal{D}_k\). The corresponding local empirical risk is
\[
J_k(\phi)
=
\frac{1}{N_k}
\sum_{i=1}^{N_k}
\mathcal{L}\bigl(P(\cdot \mid \mathbf{x}_i^k;\theta,\phi),\,y_i^k\bigr)
,
\]
where \(\mathcal{L}\) is typically the cross-entropy loss over the label space \(\mathcal{Y}\). After local optimization, node \(k\) obtains updated trainable parameters \(\phi_k\).

  \item \textbf{Aggregation:} Each node sends its updated task-specific parameters \(\phi_k\), or an equivalent update \(\Delta\phi_k\), to the server.

  \item \textbf{Global update:} The server aggregates the collection of local parameters \(\{\phi_k\}\) into a new federated global set of task-specific parameters \(\phi\). If \(M\) denotes the number of participating nodes, then a standard weighted averaging scheme can be written as
\[
\phi
\leftarrow
\sum_{k=1}^{M} \frac{N_k}{\sum_{\ell=1}^{M} N_\ell}\,\phi_k.
\]

  \item \textbf{Broadcast:} The server distributes the updated global parameters \(\phi\) back to all nodes, and the process starts again.
\end{enumerate}

Therefore, in the HFL setting for parameter-efficient LLM fine-tuning, collaboration happens via the exchange and aggregation of the trainable adaptation parameters \(\phi\), while the raw sequences \(\mathbf{x}_i^k\) and labels \(y_i^k\) remain local to each node.

\section{Centralized Datasets and Preprocessing} \label{sec:centralized-data}
In this section, we describe the datasets, outline the preprocessing steps, and present the architecture of the centralized solution to our problem.

\subsection{Description of the Datasets}
\label{sec:description}

The experiments reported in this work are conducted on four closed-ended QA and classification benchmarks spanning the financial and medical domains: FiQA-SA~\cite{maia201818}, FPB~\cite{malo2014good}, MedQA~\cite{jin2021disease}, and MedMCQA~\cite{pal2022medmcqa}. These datasets are well-suited to the evaluation of PEFT methods for LLMs, since they formulate the problem either as closed-ended question answering or as label prediction over a finite output space.

FiQA-SA is a dataset for financial sentiment analysis composed of financial news headlines and microblog posts, where each instance is annotated with a sentiment value on a continuous scale from \(-1\) (most negative) to \(1\) (most positive)~\cite{maia201818}. In practice, this benchmark evaluates whether the model can correctly infer the sentiment conveyed by short financial texts, which makes it suitable for classification-oriented fine-tuning.

FPB (Financial PhraseBank) is a benchmark for financial sentiment classification built from English financial news sentences~\cite{malo2014good}. The dataset contains 4,840 sentences annotated by multiple annotators with financial backgrounds, and each sentence is labeled according to its polarity from an investor perspective, as positive, negative, or neutral. This benchmark is particularly relevant for evaluating sentiment classification in a specialized financial language.

MedQA is a medical question answering benchmark collected from professional medical board exams~\cite{jin2021disease}. It is a multiple-choice dataset designed to assess medical knowledge and reasoning and covers three languages: English, simplified Chinese, and traditional Chinese. In the English setting, it is commonly associated with USMLE-style questions, making it a standard benchmark for closed-ended medical QA.

MedMCQA is a large-scale multi-subject medical multiple-choice question answering dataset constructed from real-world medical entrance examinations~\cite{pal2022medmcqa}. It contains more than 194,000 questions that span 21 medical topics and around 2,400 healthcare topics. Each instance includes a question, the correct answer, and distractor options, which makes it especially suitable for evaluating domain-specific reasoning and answer selection.

Together, these four datasets provide a diverse but coherent evaluation setting for LLM fine-tuning. FiQA-SA and FPB evaluate financial sentiment classification, whereas MedQA and MedMCQA assess closed-ended medical question answering and reasoning.

\subsection{Preprocessing of the Datasets}
\label{sec:preprocessing-datasets}

All datasets are preprocessed for supervised fine-tuning by converting each example into a prompt-completion pair and tokenizing it with the tokenizer of the selected base model. When no padding token is available, the end-of-sequence token is used instead. The loss is computed only on the completion tokens, while the prompt and padding tokens are masked. In addition, all inputs are truncated to a maximum sequence length of \(512\) tokens, while ensuring that sufficient space remains for the target completion. 

For FiQA-SA, the official train, validation, and test splits are used. Because the original labels are continuous sentiment scores, they are mapped into three classes (negative, neutral, and positive) using thresholds at \(-0.05\) and \(0.05\), and the targets are represented as labels in \(\{A,B,C\}\).
For FPB, the official training split is randomly shuffled, and \(10\%\) of the examples are reserved for validation. The test set corresponds to the official test split. The task is defined as a three-class sentiment classification with outputs \(A\), \(B\), and \(C\).
For MedQA, the original training split is randomly shuffled, and \(10\%\) of the examples are reserved for validation. The official test split is used for the final evaluation. Each example is converted into an exam-style multiple-choice prompt, and the target is the correct answer letter. Finally, 
for MedMCQA, the original training split is randomly shuffled, and \(10\%\) of the training examples are reserved as an internal validation set for model selection. Because the dataset does not provide a publicly labeled test split, the final evaluation is conducted on the official validation split. Moreover, the dataset is filtered to retain only valid single-choice questions with non-empty answer options and a valid correct option index. Each example is then converted into an exam-style multiple-choice prompt whose target is the correct answer letter.

\subsection{Centralized Architecture}
\label{sec:cen-architecture}

\begin{figure}[ht]
    \centering
    \includegraphics[scale=0.6]{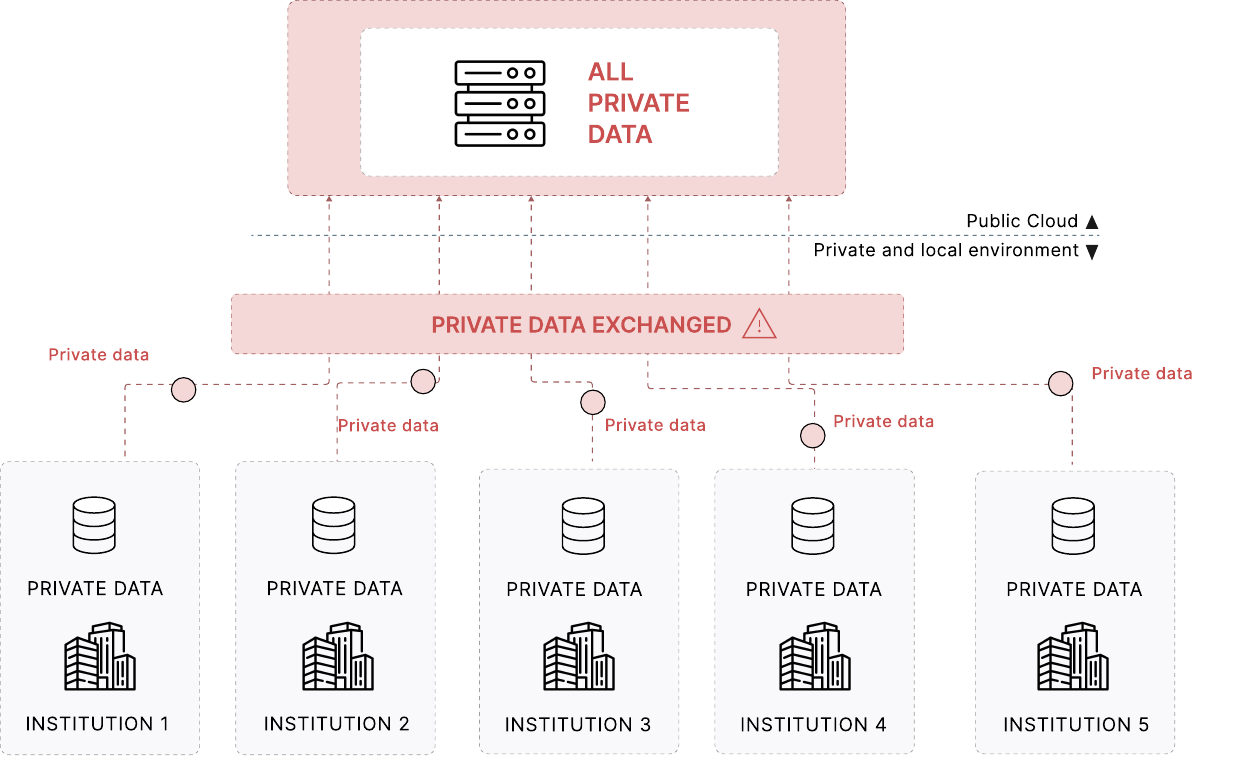}
    \caption{Classical architecture for centralized training.}
    \label{fig:cen-arch}
\end{figure}

Figure~\ref{fig:cen-arch} depicts the centralized architecture. In this setting, all local datasets are pooled into a single training repository, and model training is performed with access to the combined training data. The resulting model is therefore learned under the assumption that data from all participating institutions can be shared and processed jointly.

Given the size of the backbone model, training is performed through PEFT, where the pretrained base parameters are kept fixed or mostly frozen, and only a reduced set of task-specific parameters is optimized. This allows the model to adapt to the closed-ended QA and classification tasks while significantly reducing the number of trainable parameters and the associated computational cost.

\section{Proposed Privacy-preserving Solution through FL} \label{sec:fl-solution}

This section describes the federated benchmark setup and the procedure used to create the participating nodes.

\begin{figure}[ht]
    \centering
    \includegraphics[scale=0.54]{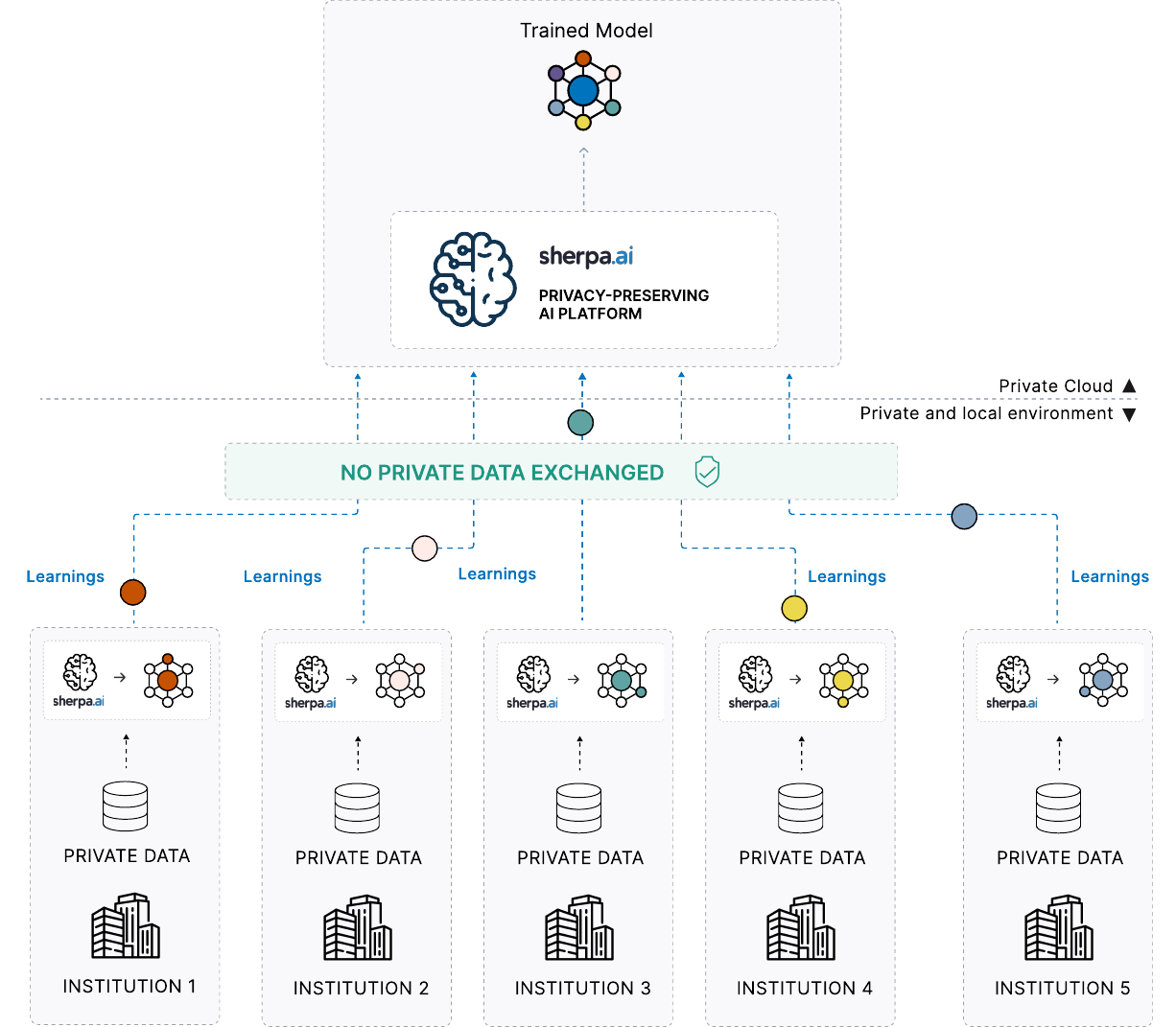}
    \caption{Proposed architecture for federated fine-tuning with privacy-preserving orchestration.}
    \label{fig:fed-arch}
\end{figure}

The node architecture used for FL is shown in Figure~\ref{fig:fed-arch}. In this scenario, raw training data are not shared across nodes; each institution keeps its local dataset on site while exchanging only trainable model parameters or updates. A central server coordinates the training process by distributing the current global task-specific parameters, collecting the locally updated parameters from the participating nodes, aggregating them, and broadcasting the updated global model back to the nodes.

Given the large size of the backbone model, training is carried out through PEFT: the pretrained base parameters remain fixed or mostly frozen, and only a reduced set of task-specific parameters is updated. This substantially decreases the computational and memory requirements of the federated process while preserving the model's ability to adapt to the target closed-ended QA and classification tasks.

\subsection{Creation of Nodes}
\label{sec:nodes_creation}

To emulate a federated setting with heterogeneous local datasets, the training split of each benchmark is partitioned across five nodes using a Dirichlet-based protocol~\cite{jimenez2024fedartml}. This procedure creates controlled label-distribution shifts across nodes, reflecting the type of statistical heterogeneity commonly observed in real-world federated environments. It also provides a reproducible protocol for evaluating federated fine-tuning under comparable non-IID conditions across all benchmarks. For the multiple-choice QA datasets, the resulting heterogeneity is defined over the target answer labels, so that different nodes receive different proportions of examples whose correct option is A, B, C, or D. More specifically, for each dataset, the partition is generated with concentration parameter \(\alpha\). Following~\cite{jimenez2024fedartml}, which uses a Hellinger distance of \(0.5\) as a reference threshold associated with noticeable non-IID effects, we select \(\alpha\) so that the resulting partitions reach at least this level of non-IID data. In particular, we set \(\alpha=0.7\) for FPB and \(\alpha=1.0\) for FiQA-SA, MedQA, and MedMCQA.

\begin{figure}[ht]
    \centering
    \begin{subfigure}{0.48\linewidth}
        \centering
        \includegraphics[width=\linewidth]{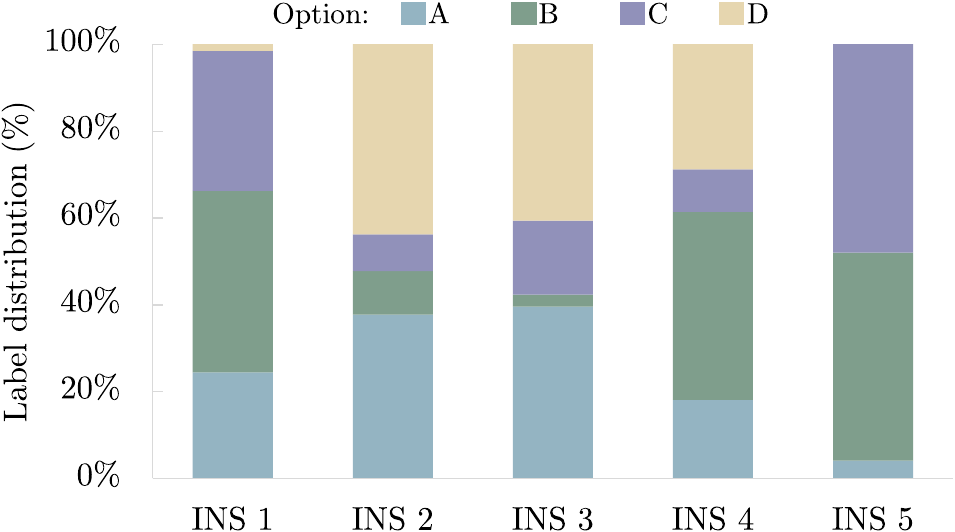}
        \caption{MedQA}
        \label{fig:distro_medqa}
    \end{subfigure}\hfill
    \begin{subfigure}{0.48\linewidth}
        \centering
        \includegraphics[width=\linewidth]{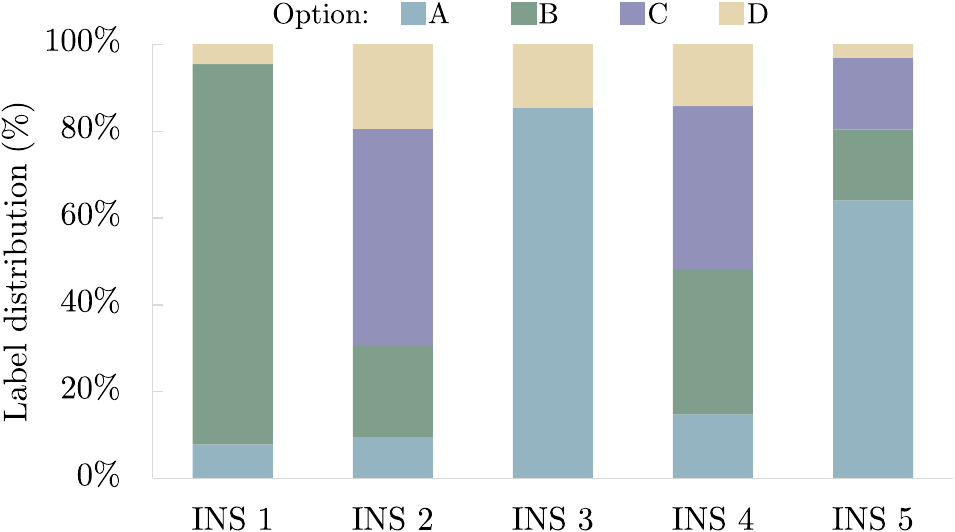}
        \caption{MedMCQA}
        \label{fig:distro_medmcqa}
    \end{subfigure}

    \vspace{0.2cm}

    \begin{subfigure}{0.48\linewidth}
        \centering
        \includegraphics[width=\linewidth]{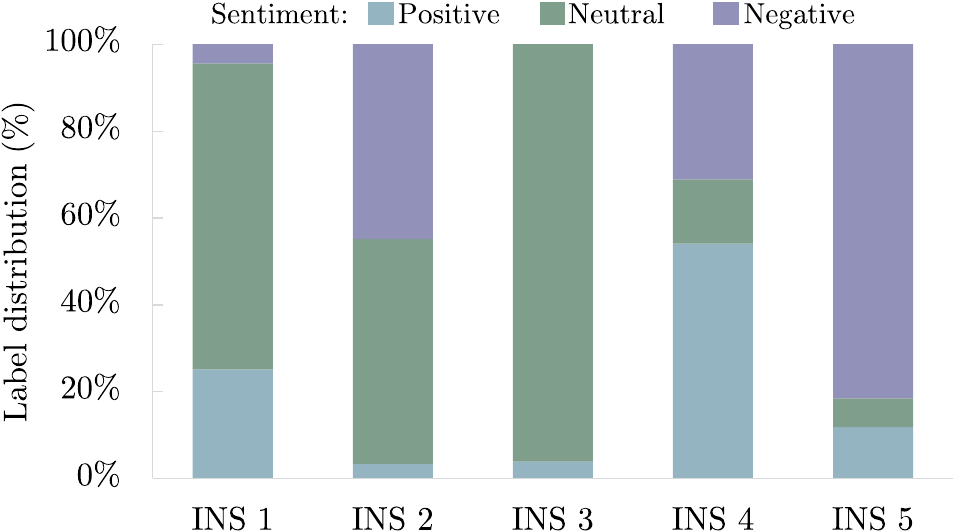}
        \caption{FPB}
        \label{fig:distro_fpb}
    \end{subfigure}\hfill
    \begin{subfigure}{0.48\linewidth}
        \centering
        \includegraphics[width=\linewidth]{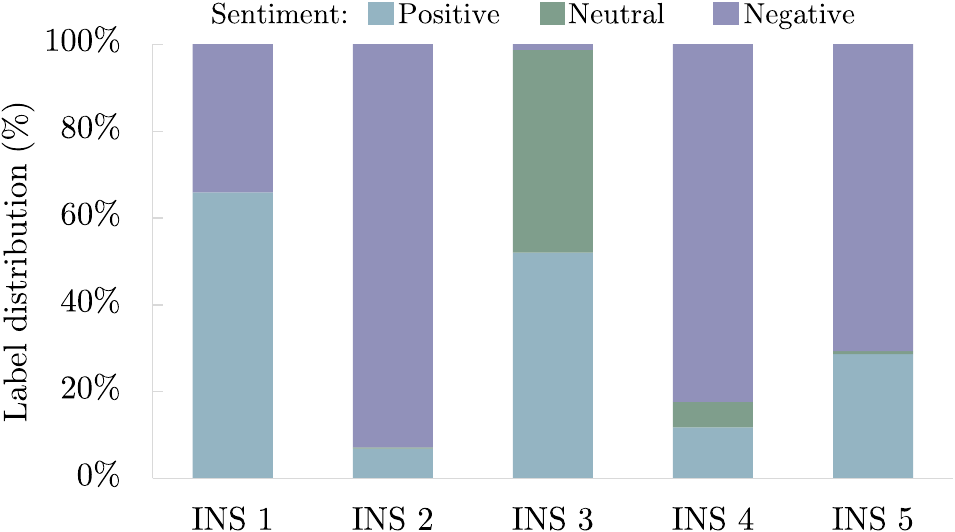}
        \caption{FiQA-SA}
        \label{fig:distro_fiqasa}
    \end{subfigure}

    \caption{Label distribution across institutions (INS) for the non-IID partitions used in each dataset. Each stacked bar shows the proportion of class labels within one institution.}
    \label{fig:data_distribution_per_node}
\end{figure}

Figure~\ref{fig:data_distribution_per_node} illustrates the resulting label distributions across the five institutions for the four datasets. The plots show that the Dirichlet-based partitioning induces heterogeneous class proportions across institutions. For the medical QA datasets, the labels correspond to correct answer options, whereas for the financial classification datasets, they correspond to sentiment categories.

\section{Experiments}
\label{sec:experiments}

In this section, we describe the experimental setup, including the evaluation metric and the training configurations, and present the main results.

We evaluate the models under three training scenarios on FiQA-SA, FPB, MedQA, and MedMCQA:

\begin{enumerate}
    \item \textbf{Centralized}: All training data are pooled into a single centralized repository, and the model is fine-tuned using the complete training split. Evaluation is then performed on the corresponding evaluation set of each dataset.
    
    \item \textbf{Federated}: The training split of each dataset is partitioned across multiple nodes according to the Dirichlet-based protocol described in Section~\ref{sec:fl-solution}. Each node performs local fine-tuning on its own private data, and the local updates are aggregated by the central server over successive communication rounds. The resulting global model is evaluated on the corresponding evaluation set.
    
    \item \textbf{Single-institution:} Each node fine-tunes its own model independently using only its local partition of the training data, without any collaboration or parameter aggregation. Each locally trained model is evaluated on the corresponding evaluation set. For each model, PEFT method, and dataset, the single-institution value reported in the tables corresponds to the mean of the accuracies over the five different nodes.
\end{enumerate}

These three scenarios enable a direct comparison between full data centralization, fully isolated local training, and collaborative training under data-locality constraints. In this way, the experiments quantify the extent to which FL provides an effective compromise between avoiding raw data sharing and still benefiting from the distributed knowledge available across nodes.

\subsection{Evaluation Metrics}
\label{sec:eval_metrics}

We evaluate predictive performance using accuracy, a standard metric for closed-ended question answering and classification tasks. It measures the proportion of examples for which the predicted output exactly matches the ground-truth label.

Formally, accuracy is defined as
\begin{align}
    \text{Accuracy} = \frac{1}{N}\sum_{i=1}^{N}\mathbb{I}\bigl(\hat{y}_i = y_i\bigr),
\end{align}
where \(N\) is the total number of evaluated examples, \(y_i\) is the ground-truth label, \(\hat{y}_i\) is the predicted label, and \(\mathbb{I}(\cdot)\) is the indicator function, which takes value \(1\) if its argument is true and \(0\) otherwise.

In the federated setting, we also report the communication cost per round and the memory footprint. The communication cost per round is defined as the bidirectional transmitted volume of trainable parameters between the server and the participating nodes in a single communication round:
\begin{align}
    \mathrm{\text{Communication cost per round}}
    =
    \frac{2 K B_{\mathrm{train}}}{1024^3},
\end{align}
where \(K\) is the number of participating nodes per round and \(B_{\mathrm{train}}\) is the number of communicated trainable bytes per client.

The memory footprint is defined as the mean peak reserved GPU memory as reported by PyTorch across participating nodes:
\begin{align}
    \mathrm{\text{Memory footprint}}
    =
    \frac{1}{K}\sum_{k=1}^{K} \frac{M_k}{1024^3},
\end{align}
where \(M_k\) denotes the peak reserved GPU memory in bytes observed for node \(k\) during training.


\subsection{Reproducibility Details}
\label{sec:reproducibility-details}

All experiments are carried out on a machine with 219 GB of disk storage, an AMD EPYC 7R13 processor with 48 cores, 372 GB of RAM, Debian GNU/Linux 13 (Trixie), and four NVIDIA L40S GPUs, each with 46,068 MB of memory and Python~3.11. All models are trained using the \sherpa FL platform.

Unless otherwise stated, all experiments use a fixed random seed of \(42\). The optimization process uses the $paged\_adamw\_8bit$ optimizer, a learning rate of \(10^{-4}\), gradient accumulation of \(1\), and a per-device batch size of \(4\). The maximum input length is \(512\) tokens. Gradient checkpoints are enabled, the maximum gradient norm is set to \(1.0\), and the learning-rate scheduler is constant with \(10\) warmup steps. Mixed-precision training is used whenever supported by the hardware, with $bfloat16$ preferred over $float16$.

For the PEFT configuration, LoRA and QLoRA use rank \(r=32\), scaling factor \(\alpha=21\), and dropout \(0.1\). Their target modules are inferred automatically from the model architecture at runtime. For the adapter-based setting, we use IA3 adapters applied to the modules $k\_proj$, $v\_proj$, and $down\_proj$, with $down\_proj$ specified as the feedforward module.

In the federated setup, all available nodes participate in every communication round, and model aggregation is performed with FedAvg over \(4\) communication rounds, using \(1\) local epoch per round. Thus, each node performs a total of \(4\) local epochs in the federated scenario. To ensure a fair comparison across training settings, the centralized and single-institution baselines are also trained for \(4\) epochs.

The models employed in this work are II-Medical-8B-1706~\cite{2025II-Medical-8B-1706}, Qwen/Qwen3-8B~\cite{qwen3technicalreport}, Qwen/Qwen3-4B-Instruct-2507~\cite{qwen3technicalreport}, Qwen/Qwen2.5-1.5B-Instruct~\cite{qwen2.5}, meta-llama/Llama-3.1-8B-Instruct~\cite{grattafiori2024llama3herd}, meta-llama/Meta-Llama-3-8B-Instruct~\cite{llama3modelcard}, hongzhouyu/FineMedLM-o1~\cite{yu2025finemedlm}, google/gemma-2-9b-it~\cite{gemma_2024}, and google/gemma-3-4b-it~\cite{gemma_2025}. These models were selected because they either showed strong performance in prior studies or correspond to recent versions of model families that have consistently achieved competitive results.

\subsection{Results}

Table~\ref{tab:benchmark-medical} summarizes our experimental results for the medical datasets. It shows a clear advantage of the medical-specialized backbone II-Medical-8B-1706, which achieves the best accuracy in all three training scenarios and for both medical benchmarks. In particular, its LoRA configuration reaches \(0.740\) and \(0.763\) in the centralized setting for MedQA and MedMCQA, respectively, while also obtaining the strongest federated results, namely \(0.714\) and \(0.742\). This suggests that domain specialization is particularly beneficial in the medical setting, where models explicitly designed for clinical reasoning transfer more effectively across PEFT strategies and training scenarios.

\begin{table}[ht]
  \scriptsize
  \setlength{\tabcolsep}{4pt}
  \renewcommand{\arraystretch}{1.25}
  \centering
  \begin{tabularx}{\textwidth}{c|c|YYY|YYY}
    \toprule
    \multirow{2}{*}{\textbf{Model}} & \multirow{2}{*}{\textbf{PEFT}} &
    \multicolumn{3}{c|}{\textbf{MedQA}} &
    \multicolumn{3}{c}{\textbf{MedMCQA}} \\
    \cmidrule(lr){3-5}\cmidrule(lr){6-8}
    & & \textbf{Single-institution} & \textbf{Centralized} & \textbf{Federated}
      & \textbf{Single-institution} & \textbf{Centralized} & \textbf{Federated} \\
    \midrule
    \multirow{3}{*}{II-Medical-8B-1706}
      & LoRA     & \textbf{0.659} & \textbf{0.740} & \textbf{0.714} & \textbf{0.593} & \textbf{0.763} & \textbf{0.742} \\
      & QLoRA    & 0.650 & 0.725 & 0.686 & 0.582 & 0.755 & 0.741 \\
      & IA3 Adapters & 0.657 & 0.705 & 0.692 & 0.571 & 0.735 & 0.694 \\
    \midrule
    \multirow{3}{*}{Qwen3-8B}
      & LoRA     & 0.554 & 0.696 & 0.683 & 0.547 & 0.746 & 0.718 \\
      & QLoRA    & 0.530 & 0.663 & 0.669 & 0.543 & 0.720 & 0.683 \\
      & IA3 Adapters & 0.550 & 0.655 & 0.629 & 0.493 & 0.694 & 0.656 \\
    \midrule
    \multirow{3}{*}{Qwen3-4B-Instruct-2507}
      & LoRA     & 0.562 & 0.650 & 0.646 & 0.523 & 0.704 & 0.686 \\
      & QLoRA    & 0.546 & 0.644 & 0.629 & 0.519 & 0.650 & 0.672 \\
      & IA3 Adapters & 0.555 & 0.627 & 0.612 & 0.493 & 0.698 & 0.633 \\
    \midrule
    \multirow{3}{*}{Qwen2.5-1.5B-Instruct}
      & LoRA     & 0.427 & 0.512 & 0.532 & 0.439 & 0.620 & 0.592 \\
      & QLoRA    & 0.403 & 0.480 & 0.500 & 0.433 & 0.586 & 0.564 \\
      & IA3 Adapters & 0.383 & 0.486 & 0.468 & 0.379 & 0.556 & 0.528 \\
    \midrule
    \multirow{3}{*}{Llama-3.1-8B-Instruct}
      & LoRA     & 0.579 & 0.656 & 0.651 & 0.547 & 0.687 & 0.710 \\
      & QLoRA    & 0.567 & 0.634 & 0.639 & 0.531 & 0.692 & 0.684 \\
      & IA3 Adapters & 0.616 & 0.635 & 0.638 & 0.436 & 0.665 & 0.622 \\
    \midrule
    \multirow{3}{*}{Meta-Llama-3-8B-Instruct}
      & LoRA     & 0.548 & 0.633 & 0.637 & 0.542 & 0.669 & 0.712 \\
      & QLoRA    & 0.532 & 0.628 & 0.622 & 0.554 & 0.680 & 0.689 \\
      & IA3 Adapters & 0.588 & 0.615 & 0.616 & 0.534 & 0.697 & 0.690 \\
    \midrule
    \multirow{3}{*}{FineMedLM-o1}
      & LoRA     & 0.476 & 0.634 & 0.611 & 0.467 & 0.646 & 0.527 \\
      & QLoRA    & 0.467 & 0.613 & 0.589 & 0.458 & 0.667 & 0.642 \\
      & IA3 Adapters & 0.474 & 0.549 & 0.528 & 0.351 & 0.542 & 0.527 \\
    \midrule
    \multirow{3}{*}{gemma-2-9b-it}
      & LoRA     & 0.568 & 0.675 & 0.665 & 0.515 & 0.703 & 0.691 \\
      & QLoRA    & 0.541 & 0.677 & 0.653 & 0.492 & 0.714 & 0.693 \\
      & IA3 Adapters & 0.594 & 0.645 & 0.634 & 0.510 & 0.665 & 0.657 \\
    \midrule
    \multirow{3}{*}{gemma-3-4b-it}
      & LoRA     & 0.421 & 0.533 & 0.529 & 0.437 & 0.608 & 0.573 \\
      & QLoRA    & 0.397 & 0.521 & 0.502 & 0.429 & 0.537 & 0.575 \\
      & IA3 Adapters & 0.408 & 0.460 & 0.473 & 0.394 & 0.555 & 0.579 \\
    \bottomrule
  \end{tabularx}
  \vspace{4mm}
  \caption{Accuracy comparison across models, fine-tuning methods, and training scenarios in the \textbf{medical} domain. Results are reported for Single-institution, Centralized, and Federated scenarios. The highest performance is indicated in \textbf{bold}.}
  \label{tab:benchmark-medical}
\end{table}

A second relevant trend is that federated training often remains close to centralized training while improving over the single-institution setting in most cases. This pattern is especially visible for stronger backbones such as Qwen3-8B, Llama-3.1-8B-Instruct, and Meta-Llama-3-8B-Instruct, where federated training preserves a substantial portion of the centralized performance. Therefore, FL appears to provide an effective balance between isolated local training and full data centralization in the medical domain across many of the evaluated configurations.

Regarding PEFT methods, LoRA is generally the strongest option in the medical benchmarks, with QLoRA remaining a close second in most cases, and IA3 adapters usually trailing behind. The gap is not substantial for the strongest models, but it becomes more visible for smaller or less specialized backbones, which indicates that LoRA has the most robust accuracy profile in the domain.

\begin{figure}[ht]
    \centering
    \begin{subfigure}{0.48\linewidth}
        \centering
        \includegraphics[width=\linewidth]{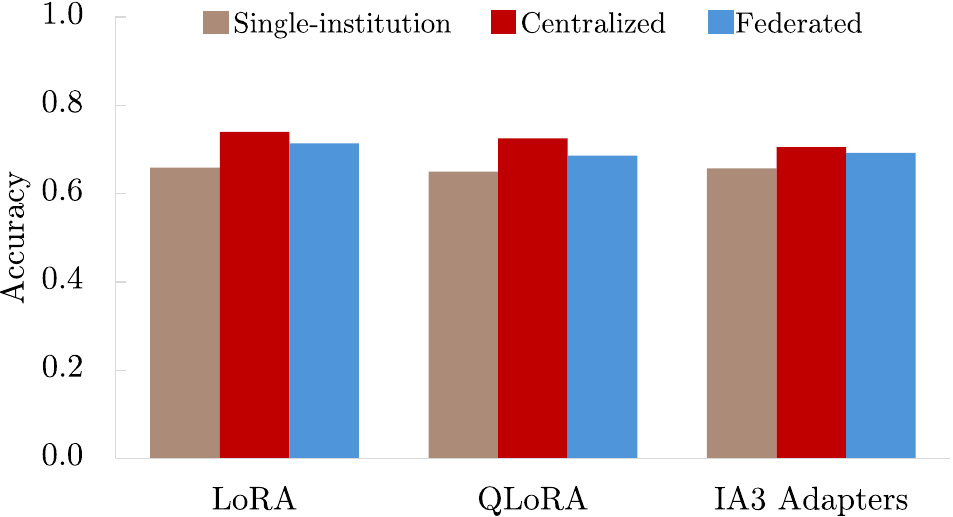}
        \caption{MedQA}
        \label{fig:accuracy_medqa_best_model}
    \end{subfigure}\hfill
    \begin{subfigure}{0.48\linewidth}
        \centering
        \includegraphics[width=\linewidth]{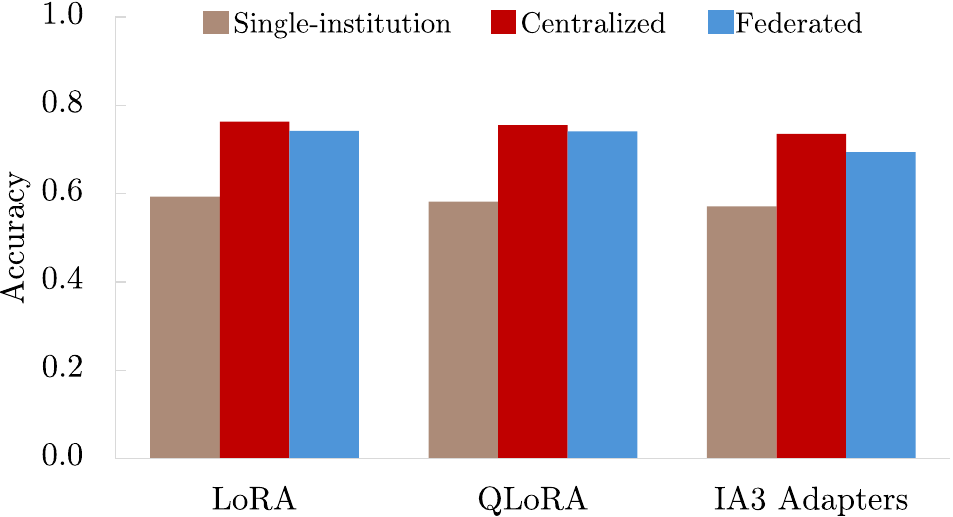}
        \caption{MedMCQA}
        \label{fig:accuracy_medmcqa_best_model}
    \end{subfigure}
    \caption{Accuracy for the Single-institution, Centralized, and Federated scenarios for the best model for the \textbf{medical} domain (II-Medical-8B-1706).}
    \label{fig:acc_best_model_medical}
\end{figure}

Figure~\ref{fig:acc_best_model_medical} provides a focused view of II-Medical-8B-1706, the strongest medical backbone in our experiments. The figure illustrates the same trend observed in Table~\ref{tab:benchmark-medical}: centralized training usually achieves the highest accuracy, federated training remains close, and single-institution training is consistently lower. This confirms that the federated setting captures useful cross-node information even under non-IID partitions.

The figure also highlights that LoRA is the most effective PEFT method for this model, while QLoRA provides nearly the same predictive quality with a small reduction in accuracy. IA3 adapters remain competitive, but their performance is systematically lower than that of LoRA and QLoRA, particularly on MedMCQA. This indicates that, for highly specialized medical models, LoRA offers the best accuracy, whereas QLoRA may be preferred when memory efficiency is a priority. 

Table~\ref{tab:benchmark-financial} summarizes our experiments on the financial datasets. The results are more heterogeneous than those of the medical domain, since no single model dominates all metrics and scenarios. In the centralized setting, the strongest results are obtained by Qwen2.5-1.5B-Instruct on FPB with LoRA (\(0.873\)) and by Qwen3-8B on FiQA-SA with LoRA (\(0.915\)). In the federated setting, the best FPB result is achieved by Qwen3-8B (\(0.863\)), while the best FiQA-SA is obtained by gemma-2-9b-it with LoRA (\(0.923\)). These results indicate that financial-domain performance is more task-dependent and that the preferred backbone may vary across benchmarks.

\begin{table}[ht]
  \scriptsize
  \setlength{\tabcolsep}{4pt}
  \renewcommand{\arraystretch}{1.25}
  \centering
  \begin{tabularx}{\textwidth}{c|c|YYY|YYY}
    \toprule
    \multirow{2}{*}{\textbf{Model}} & \multirow{2}{*}{\textbf{PEFT}} &
    \multicolumn{3}{c|}{\textbf{FPB}} &
    \multicolumn{3}{c}{\textbf{FiQA-SA}} \\
    \cmidrule(lr){3-5}\cmidrule(lr){6-8}
    & & \textbf{Single-institution} & \textbf{Centralized} & \textbf{Federated}
      & \textbf{Single-institution} & \textbf{Centralized} & \textbf{Federated} \\
    \midrule
    \multirow{3}{*}{Qwen3-8B}
      & LoRA     & 0.691 & 0.872 & \textbf{0.863} & 0.650 & \textbf{0.915} & 0.902 \\
      & QLoRA    & 0.664 & 0.868 & 0.863 & 0.634 & 0.897 & 0.910 \\
      & IA3 Adapters & 0.437 & 0.809 & 0.809 & 0.416 & 0.808 & 0.822 \\
    \midrule
    \multirow{3}{*}{Qwen3-4B-Instruct-2507}
      & LoRA     & 0.636 & 0.870 & 0.847 & 0.655 & 0.897 & 0.885 \\
      & QLoRA    & 0.678 & 0.869 & 0.842 & 0.740 & 0.880 & 0.876 \\
      & IA3 Adapters & 0.670 & 0.789 & 0.779 & 0.854 & 0.863 & 0.859 \\
    \midrule
    \multirow{3}{*}{Qwen2.5-1.5B-Instruct}
      & LoRA     & 0.637 & \textbf{0.873} & 0.831 & 0.684 & 0.859 & 0.859 \\
      & QLoRA    & 0.580 & 0.851 & 0.832 & 0.679 & 0.855 & 0.838 \\
      & IA3 Adapters & 0.663 & 0.764 & 0.762 & 0.803 & 0.803 & 0.829 \\
    \midrule
    \multirow{3}{*}{Llama-3.1-8B-Instruct}
      & LoRA     & 0.612 & 0.866 & 0.846 & 0.771 & 0.897 & 0.910 \\
      & QLoRA    & 0.654 & 0.857 & 0.846 & 0.691 & 0.902 & 0.889 \\
      & IA3 Adapters & 0.628 & 0.776 & 0.758 & 0.718 & 0.846 & 0.880 \\
    \midrule
    \multirow{3}{*}{Meta-Llama-3-8B-Instruct}
      & LoRA     & 0.621 & 0.868 & 0.841 & 0.718 & 0.902 & 0.910 \\
      & QLoRA    & 0.641 & 0.864 & 0.845 & 0.700 & 0.910 & 0.902 \\
      & IA3 Adapters & 0.696 & 0.774 & 0.765 & 0.735 & 0.799 & 0.812 \\
    \midrule
    \multirow{3}{*}{gemma-2-9b-it}
      & LoRA     & 0.661 & 0.857 & 0.852 & 0.702 & 0.906 & \textbf{0.923} \\
      & QLoRA    & 0.604 & 0.871 & 0.861 & 0.764 & 0.906 & 0.915 \\
      & IA3 Adapters & \textbf{0.746} & 0.808 & 0.805 & 0.822 & 0.850 & 0.889 \\
    \midrule
    \multirow{3}{*}{gemma-3-4b-it}
      & LoRA     & 0.576 & 0.865 & 0.844 & 0.711 & 0.863   & 0.880 \\
      & QLoRA    & 0.612 & 0.859 & 0.835 & 0.697 & 0.859   & 0.872 \\
      & IA3 Adapters & 0.677 & 0.763 & 0.732 & \textbf{0.856} & 0.842   & 0.868 \\
    \bottomrule
  \end{tabularx}
  \vspace{4mm}
  \caption{Accuracy comparison across models, fine-tuning methods, and training scenarios in the \textbf{financial} domain. Results are reported for Single-institution, Centralized, and Federated scenarios. The highest performance is indicated in \textbf{bold}.}
  \label{tab:benchmark-financial}
\end{table}

Similar to the medical domain, federated training is close to centralized training in most configurations, while it often outperforms single-institution training. In particular, the federated results remain competitive for several models, especially Qwen3-8B, Llama-3.1-8B-Instruct, and gemma-2-9b-it. These results show that FL can preserve a large fraction of the predictive performance even in financial tasks with highly non-IID node distributions.

In terms of PEFT, LoRA again provides the most stable overall performance, although QLoRA remains highly competitive and in some cases nearly matches the LoRA results. IA3 adapters are less consistent: they are clearly weaker on some settings, but they also yield the best single-institution results for FPB and FiQA-SA with gemma-2-9b-it and gemma-3-4b-it. This suggests that the relative effectiveness of PEFT methods in finance depends more on the backbone and the dataset when compared to their effectiveness in the medical domain. 

\begin{figure}[ht]
    \centering
    \begin{subfigure}{0.48\linewidth}
        \centering
        \includegraphics[width=\linewidth]{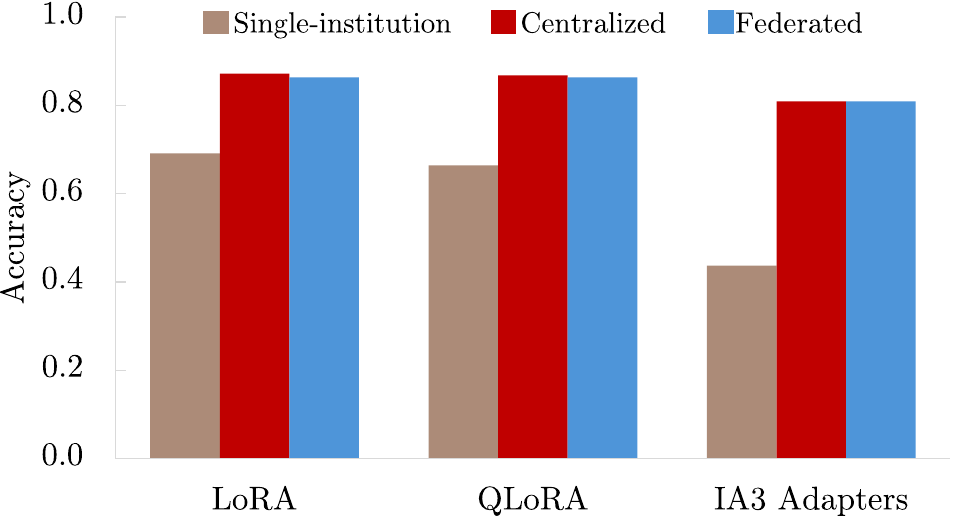}
        \caption{FPB}
        \label{fig:accuracy_fpb_best_model}
    \end{subfigure}\hfill
    \begin{subfigure}{0.48\linewidth}
        \centering
        \includegraphics[width=\linewidth]{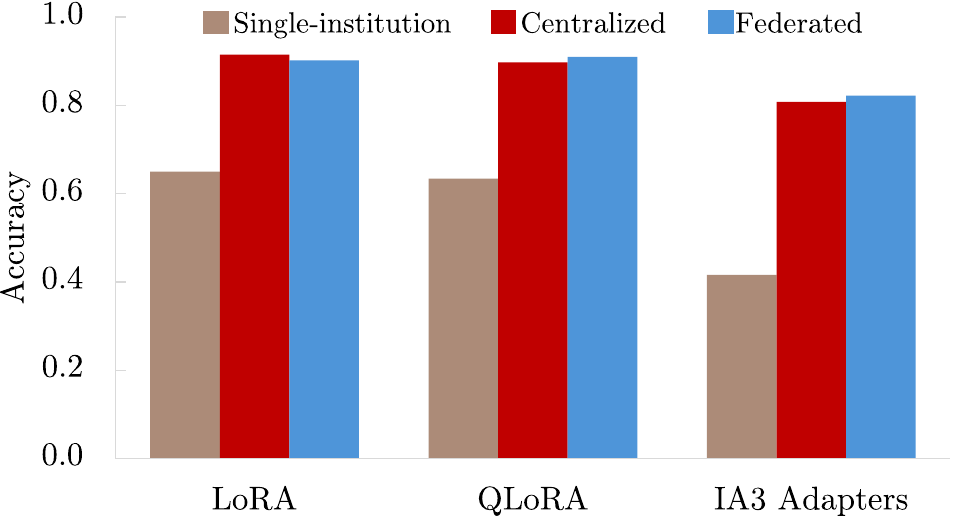}
        \caption{FiQA-SA}
        \label{fig:accuracy_fiqasa_best_model}
    \end{subfigure}
    \caption{Accuracy for the Single-institution, Centralized, and Federated scenarios for a representative \textbf{financial-domain} model (Qwen3-8B).}
    \label{fig:acc_best_model_financial}
\end{figure}

Figure~\ref{fig:acc_best_model_financial} illustrates the behavior of a representative financial domain model, namely Qwen3-8B. For both FPB and FiQA-SA, federated training remains close to centralized training, sometimes matching or slightly exceeding it, while outperforming the single-institution setting. This confirms that the collaborative setting is beneficial in finance as well, even though the absolute gains over single-institution training are somewhat benchmark-dependent. 

The figure also shows that LoRA and QLoRA perform similarly for Qwen3-8B, especially in the federated scenario, where their FPB results are identical, and the FiQA-SA difference is small. IA3 adapters, in contrast, lead to a more pronounced drop in accuracy. Therefore, for this backbone, QLoRA appears particularly attractive because it preserves most of the predictive power of LoRA while offering a substantially lower memory footprint, as shown in Table~\ref{tab:cost-memory}.

\begin{table}[h]
  \scriptsize
  \setlength{\tabcolsep}{4pt}
  \renewcommand{\arraystretch}{1.25}
  \centering
  \begin{tabularx}{\textwidth}{c|c|c|YY|YY}
    \toprule
    \multirow{2}{*}{\textbf{Model}} &
    \multirow{2}{*}{\textbf{PEFT}} &
    \multirow{2}{*}{\textbf{Communication cost per round (GB)}} &
    \multicolumn{2}{c|}{\textbf{Memory footprint medical (GB)}} &
    \multicolumn{2}{c}{\textbf{Memory footprint financial (GB)}} \\
    \cmidrule(lr){4-5}\cmidrule(lr){6-7}
    & & & \textbf{MedQA} & \textbf{MedMCQA} & \textbf{FPB} & \textbf{FiQA-SA} \\
    \midrule
    \multirow{3}{*}{II-Medical-8B-1706}
      & LoRA     & 1.510 & 35.878 & 35.896 & NA     & NA     \\
      & QLoRA    & 1.510 & 24.275 & 24.177 & NA     & NA     \\
      & IA3 Adapters & 0.012 & 35.135 & 35.137 & NA     & NA     \\
    \midrule
    \multirow{3}{*}{Qwen3-8B}
      & LoRA     & 1.510 & 28.712 & 28.713 & 28.713 & 28.712 \\
      & QLoRA    & 1.510 & 20.750 & 20.756 & 20.570 & 20.616 \\
      & IA3 Adapters & 0.018 & 28.519 & 28.520 & 28.518 & 28.517 \\
    \midrule
    \multirow{3}{*}{Qwen3-4B-Instruct-2507}
      & LoRA     & 0.944 & 16.940 & 16.948 & 16.952 & 16.952 \\
      & QLoRA    & 0.944 & 11.756 & 11.739 & 11.778 & 11.672 \\
      & IA3 Adapters & 0.011 & 16.481 & 16.488 & 16.489 & 16.488 \\
    \midrule
    \multirow{3}{*}{Qwen2.5-1.5B-Instruct}
      & LoRA     & 0.440 & 9.519  & 10.872 & 9.524  & 9.523 \\
      & QLoRA    & 0.440 & 8.521  & 8.582  & 8.536  & 8.598 \\
      & IA3 Adapters & 0.005 & 9.419  & 9.423  & 9.424  & 9.424 \\
    \midrule
    \multirow{3}{*}{Llama-3.1-8B-Instruct}
      & LoRA     & 1.342 & 27.903 & 29.151 & 27.914 & 27.915 \\
      & QLoRA    & 1.342 & 18.593 & 17.946 & 18.354 & 18.302 \\
      & IA3 Adapters & 0.016 & 27.403 & 27.405 & 27.402 & 27.401 \\
    \midrule
    \multirow{3}{*}{Meta-Llama-3-8B-Instruct}
      & LoRA     & 1.342 & 27.903 & 27.910 & 27.914 & 27.915 \\
      & QLoRA    & 1.342 & 18.339 & 18.117 & 18.097 & 17.967 \\
      & IA3 Adapters & 0.025 & 27.403 & 27.405 & 27.402 & 27.401 \\
    \midrule
    \multirow{3}{*}{FineMedLM-o1}
      & LoRA     & 1.340 & 34.490 & 34.505 & NA     & NA     \\
      & QLoRA    & 1.340 & 21.371 & 21.334 & NA     & NA     \\
      & IA3 Adapters & 0.020 & 33.883 & 33.885 & NA     & NA     \\
    \midrule
    \multirow{3}{*}{gemma-2-9b-it}
      & LoRA     & 1.541 & 43.499 & 43.504 & 43.506 & 43.506 \\
      & QLoRA    & 1.541 & 27.915 & 27.581 & 26.977 & 27.984 \\
      & IA3 Adapters & 0.018 & 42.905 & 42.908 & 42.909 & 42.909 \\
    \midrule
    \multirow{3}{*}{gemma-3-4b-it}
      & LoRA     & 0.891 & 26.541 & 24.800 & 24.185 & 24.185 \\
      & QLoRA    & 0.891 & 21.205 & 19.457 & 18.951 & 18.856 \\
      & IA3 Adapters & 0.010 & 26.133 & 24.800 & 23.761 & 23.765 \\
    \bottomrule
  \end{tabularx}
  \vspace{4mm}
  \caption{Communication cost per round and memory footprint across models and fine-tuning methods for the Federated scenario. The lowest system overhead is indicated in \textbf{bold}. NA denotes not applicable to that domain.}
  \label{tab:cost-memory}
\end{table}

Table~\ref{tab:cost-memory} reports the communication cost and memory footprint of federated training. It highlights a clear trade-off between predictive performance and system efficiency. First, LoRA and QLoRA exhibit the same communication cost per round for a given model, which is expected because both methods communicate task-specific low-rank updates of similar size. In contrast, IA3 adapters consistently require much less communication, often by two orders of magnitude, making them the most communication-efficient option in the federated setting.

Moreover, QLoRA is systematically the most memory-efficient among the high-performing PEFT methods. For example, with Qwen3-8B, GPU memory decreases from approximately 28.7~GB under LoRA to 20.7~GB under QLoRA, with similar reductions across the remaining backbones. This confirms that quantization substantially reduces resource requirements while preserving competitive accuracy, making QLoRA particularly suitable for resource-constrained federated deployments.

Finally, IA3 adapters and LoRA tend to have relatively similar memory footprints, even though their communication costs differ substantially. This is consistent with the fact that, in both cases, the frozen backbone remains the dominant contributor to the total memory usage, whereas the size of the trainable PEFT module only has a secondary effect on the overall footprint. Consequently, from a systems perspective, IA3 adapters are primarily advantageous for reducing communication, whereas QLoRA is a better option for reducing memory.
\begin{figure}[H]
    \centering
    \begin{subfigure}{0.48\linewidth}
        \centering
        \includegraphics[width=\linewidth]{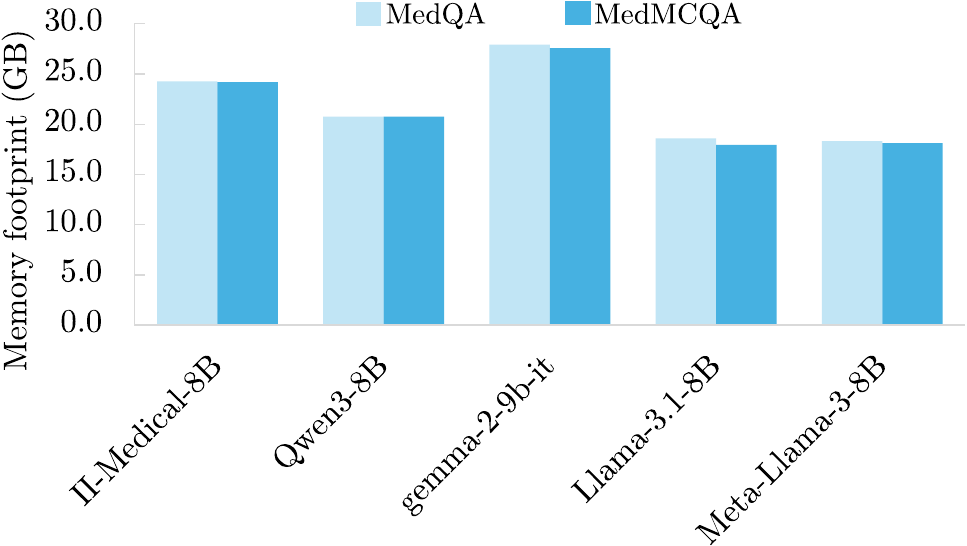}
        \caption{Medical}
        \label{fig:mem_footprint_medical}
    \end{subfigure}\hfill
    \begin{subfigure}{0.48\linewidth}
        \centering
        \includegraphics[width=\linewidth]{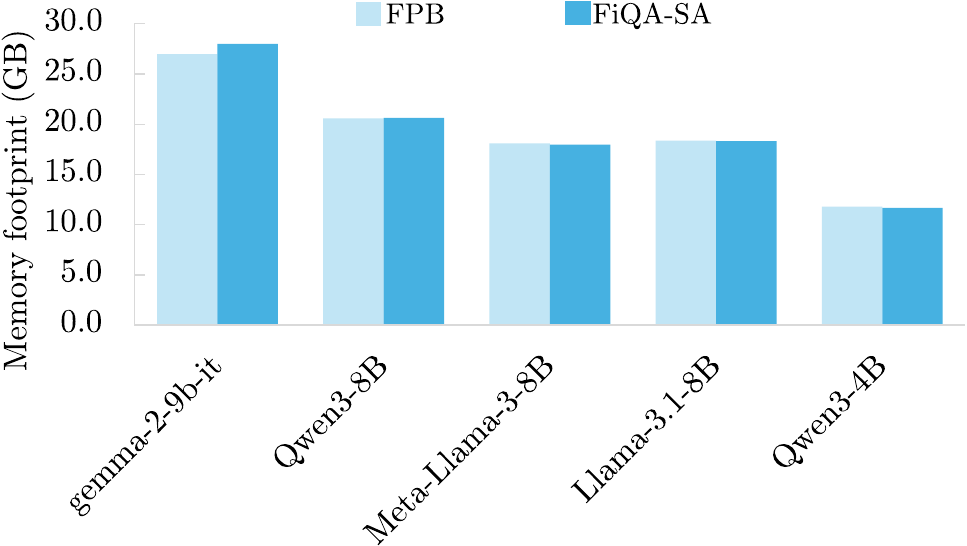}
        \caption{Financial}
        \label{fig:mem_footprint_financial}
    \end{subfigure}
    \caption{Memory footprint (GB) in the Federated scenario for the five best-performing models under QLoRA.}
    \label{fig:mem_footprint}
\end{figure}
Figure~\ref{fig:mem_footprint} focuses on the memory footprint of the five best-performing QLoRA models in each domain. In the medical setting, the best model, II-Medical-8B-1706, also incurs the highest memory consumption, whereas Qwen3-8B, Llama-3.1-8B-Instruct, and Meta-Llama-3-8B-Instruct provide a balance between accuracy and memory usage. In particular, Qwen3-4B-Instruct-2507 offers the lowest memory footprint among the top medical candidates, although at a moderate cost in predictive performance. 

In the financial setting, we can see a similar pattern. Gemma-2-9b-it is among the strongest models in terms of accuracy, but it is also the most memory-demanding. By contrast, Qwen3-8B, Llama-3.1-8B-Instruct, and Meta-Llama-3-8B-Instruct provide competitive performance at noticeably lower memory cost, while Qwen3-4B-Instruct-2507 stands out again as the most lightweight alternative among the top-performing models. Overall, the figure suggests that QLoRA enables a meaningful reduction in memory footprint, but the final model choice still depends on the desired trade-off between efficiency and predictive quality.

\section{Discussion}
\label{sec:discussion}

The results presented in Section~\ref{sec:experiments} reveal several consistent patterns across domains, training scenarios, and PEFT methods. In particular, the federated setting achieves performance close to the centralized setting in most cases, often reproducing a large fraction of its predictive performance despite operating under data locality constraints. At the same time, it tends to outperform the single-institution scenario, showing that collaborative fine-tuning across distributed nodes can effectively exploit complementary information without requiring raw data sharing.

A second important observation is that the most suitable model depends on the deployment objective. When the main goal is to maximize predictive performance in the medical domain, II-Medical-8B-1706 is the best option, as it achieves the best results across MedQA and MedMCQA in all training scenarios. In contrast, for the financial domain, the results are more heterogeneous, with Qwen3-8B, Qwen2.5-1.5B-Instruct, and gemma-2-9b-it excelling on different benchmarks and scenarios. Therefore, if the priority is raw predictive accuracy, domain-specialized models appear preferable in medicine, whereas in finance, the best choice is more task-oriented.

From a practical perspective, the preferred PEFT strategy depends on the deployment constraint. When the objective is maximum accuracy and computational cost is secondary, LoRA is generally the most reliable choice, as it provides strong and stable performance across most evaluated settings. When memory is the main constraint, QLoRA offers the best trade-off, consistently reducing memory usage while remaining close to LoRA in accuracy. When communication cost is the bottleneck, IA3 adapters are particularly attractive because they substantially reduce communication per round, although often at the cost of lower predictive performance. Finally, when accuracy and efficiency are equally important, models such as Qwen3-8B, Llama-3.1-8B-Instruct, and Qwen3-4B-Instruct-2507 provide a balanced compromise between predictive quality and resource requirements.

Our medical and financial domain results are also useful when viewed alongside previously reported results on MedQA, MedMCQA, FPB, and FiQA-SA, including recent medical~\cite{liu2024large,chen2025flow,christophe2024med42,yu2025finemedlm,lobo2025impact} and financial~\cite{wang2024finlora,fatemi2024comparative,rao2025llms,wang2025finlora,gao2025flowertune} fine-tuning works, as well as reported results for strong proprietary or domain-specific systems such as GPT-4 and Med-PaLM~2 in the medical leaderboard~\cite{medicalllmleaderboard2024}. However, these references should be interpreted only as contextual points of comparison, not as baselines that match directly, since evaluation protocols, model classes, prompting settings, data splits, and training regimes may differ across studies. This distinction is especially important because our evaluation includes federated experiments under controlled non-IID conditions, where no raw data are centralized. Therefore, rather than claiming direct superiority or equivalence, our results indicate that federated PEFT can achieve strong performance on demanding medical QA and financial classification benchmarks under data-local federated constraints.

Another important conclusion concerns the effect of non-IID data. In this study, the node partitions are explicitly controlled through a Dirichlet-based protocol with a minimum Hellinger distance threshold, which ensures a realistic but measurable degree of non-IID data. Despite this, the federated results remain close to the centralized ones in many cases, especially for stronger backbones and with LoRA or QLoRA. This suggests that, under the non-IID regimes considered here, LLM fine-tuning is affected by statistical heterogeneity but it is not severely degraded by it. In other words, non-IID data do not prevent effective federated adaptation, provided that the underlying model and PEFT strategy are sufficiently robust. This observation is aligned with recent benchmarks and systems studies in federated LLM fine-tuning, which also report that realistic client heterogeneity might be challenging but it is manageable under appropriate configurations~\cite{ye2024fedllm,gao2025flowertune,chen2025flow}.

It is also important to clarify the privacy scope of this work. FL reduces the need to centralize raw data by keeping training examples local to each node; however, this should not be interpreted as providing formal privacy guarantees by itself. Model updates may still leak information under certain threat models, for example, through gradient inversion, reconstruction attacks, or other forms of update-level inference. In this benchmark, we focus on the empirical behavior of PEFT-based LLM fine-tuning under federated, centralized, and single-institution settings, and we do not evaluate defenses against such attacks. In particular, mechanisms such as differential privacy, secure aggregation, trusted execution environments, or cryptographic protocols are outside the scope of the present study. Therefore, the results should be understood as evaluating federated fine-tuning under a data-locality assumption, rather than as demonstrating end-to-end privacy protection against adversarial leakage.

Finally, our results also give insights from a Green AI perspective. Recent work argues that the evaluation of AI systems should account not only for predictive performance but also for computational efficiency and environmental cost~\cite{schwartz2020green}. Our benchmark supports this view by showing that the choice of PEFT method can substantially alter the memory and communication requirements of federated fine-tuning. In particular, QLoRA provides a clear reduction in memory footprint with negligible loss in accuracy, while IA3 adapters sharply reduce communication overhead. These results indicate that greener LLM adaptation is not only a matter of choosing smaller models, but also of selecting the appropriate fine-tuning strategy according to the deployment constraints. Therefore, efficiency-oriented reporting should be considered as an essential component of future benchmarks for federated LLM adaptation.

\section{Conclusions} \label{sec:conclusions}

Recent advances in LLMs have largely relied on publicly available data for model training, but one of the next major leaps in their performance improvement will depend on access to private, institutionally held datasets. In sensitive and highly regulated domains such as healthcare and finance, these datasets contain critical knowledge that could substantially improve LLM performance and real-world utility. However, they are distributed across institutions and cannot be centralized due to privacy, regulatory, confidentiality, and organizational constraints. This paper demonstrated that federated PEFT provides a practical mechanism to unlock these private data silos for collaborative LLM adaptation without requiring raw data sharing.

To study this setting, we introduced a unified benchmark for PEFT of LLMs under centralized, single-institution, and federated training. The benchmark covers closed-ended medical and financial tasks under controlled non-IID conditions, using MedQA, MedMCQA, FPB, and FiQA-SA. Our results show that federated fine-tuning generally reaches performance close to centralized training while clearly outperforming isolated single-institution learning. This suggests that institutions can benefit from collaborative model adaptation even when their data remain local.

We further show that model and adapter selection play a key role in practical deployment. Domain-specialized models are especially effective in medicine, while financial tasks exhibit more task-dependent behavior. LoRA tends to achieve the highest accuracy, whereas QLoRA provides the best balance between accuracy and memory efficiency, and IA3 offers the strongest communication efficiency.

Taken together, these findings position federated PEFT as a competitive and scalable approach for data-local LLM adaptation in settings where raw data cannot be centralized. Beyond demonstrating feasibility, our results suggest that FL can help enable the next leap in LLM development by allowing models to learn from private, heterogeneous, and institutionally distributed data that would otherwise remain inaccessible.

\section*{Contributions and Acknowledgments} \label{app:A}


Daniel M. Jimenez-Gutierrez

Enrique Zuazua

Georgios Kellaris

Joaquin del Rio

Oleksii Sliusarenko

Xabi Uribe-Etxebarria

\vspace{5mm}
The authors are presented in alphabetical order by first name.

\bibliographystyle{unsrt}
\bibliography{references}

@inproceedings{mcmahan2017communication,
  title={Communication-efficient learning of deep networks from decentralized data},
  author={McMahan, Brendan and Moore, Eider and Ramage, Daniel and Hampson, Seth and y Arcas, Blaise Aguera},
  booktitle={Artificial intelligence and statistics},
  pages={1273--1282},
  year={2017},
  organization={PMLR}
}

@article{jimenez2024fedartml,
  title={FedArtML: A Tool to Facilitate the Generation of Non-IID Datasets in a Controlled Way to Support Federated Learning Research},
  author={Jimenez-Gutierrez, Daniel M and Anagnostopoulos, Aris and Chatzigiannakis, Ioannis and Vitaletti, Andrea},
  journal={IEEE Access},
  year={2024},
  publisher={IEEE}
}

@article{jimenez2024non,
  title={Non-IID data in Federated Learning: A Systematic Review with Taxonomy, Metrics, Methods, Frameworks and Future Directions},
  author={Jimenez-Gutierrez, Daniel M and Solans, David and Heikkila, Mikko and Vitaletti, Andrea and Kourtellis, Nicolas and Anagnostopoulos, Aris and Chatzigiannakis, Ioannis},
  journal={arXiv e-prints},
  pages={arXiv--2411},
  year={2024}
}

@article{acar2021federated,
  title={Federated learning based on dynamic regularization},
  author={Acar, Durmus Alp Emre and Zhao, Yue and Navarro, Ramon Matas and Mattina, Matthew and Whatmough, Paul N and Saligrama, Venkatesh},
  journal={arXiv preprint arXiv:2111.04263},
  year={2021}
}

@inproceedings{li2022federated,
  title={Federated learning on non-iid data silos: An experimental study},
  author={Li, Qinbin and Diao, Yiqun and Chen, Quan and He, Bingsheng},
  booktitle={2022 IEEE 38th international conference on data engineering (ICDE)},
  pages={965--978},
  year={2022},
  organization={IEEE}
}

@inproceedings{maia201818,
  title={Www'18 open challenge: financial opinion mining and question answering},
  author={Maia, Macedo and Handschuh, Siegfried and Freitas, Andr{\'e} and Davis, Brian and McDermott, Ross and Zarrouk, Manel and Balahur, Alexandra},
  booktitle={Companion proceedings of the the web conference 2018},
  pages={1941--1942},
  year={2018}
}

@article{malo2014good,
  title={Good debt or bad debt: Detecting semantic orientations in economic texts},
  author={Malo, Pekka and Sinha, Ankur and Korhonen, Pekka and Wallenius, Jyrki and Takala, Pyry},
  journal={Journal of the Association for Information Science and Technology},
  volume={65},
  number={4},
  pages={782--796},
  year={2014},
  publisher={Wiley Online Library}
}

@article{jin2021disease,
  title={What disease does this patient have? a large-scale open domain question answering dataset from medical exams},
  author={Jin, Di and Pan, Eileen and Oufattole, Nassim and Weng, Wei-Hung and Fang, Hanyi and Szolovits, Peter},
  journal={Applied Sciences},
  volume={11},
  number={14},
  pages={6421},
  year={2021},
  publisher={MDPI}
}

@inproceedings{pal2022medmcqa,
  title={Medmcqa: A large-scale multi-subject multi-choice dataset for medical domain question answering},
  author={Pal, Ankit and Umapathi, Logesh Kumar and Sankarasubbu, Malaikannan},
  booktitle={Conference on health, inference, and learning},
  pages={248--260},
  year={2022},
  organization={PMLR}
}

@article{christophe2024med42,
  title={Med42--evaluating fine-tuning strategies for medical LLMs: full-parameter vs. parameter-efficient approaches},
  author={Christophe, Cl{\'e}ment and Kanithi, Praveen K and Munjal, Prateek and Raha, Tathagata and Hayat, Nasir and Rajan, Ronnie and Al-Mahrooqi, Ahmed and Gupta, Avani and Salman, Muhammad Umar and Gosal, Gurpreet and others},
  journal={arXiv preprint arXiv:2404.14779},
  year={2024}
}

@article{fatemi2024comparative,
author = {Fatemi, Sorouralsadat and Hu, Yuheng and Mousavi, Maryam},
title = {A Comparative Analysis of Instruction Fine-Tuning Large Language Models for Financial Text Classification},
year = {2025},
issue_date = {March 2025},
publisher = {Association for Computing Machinery},
address = {New York, NY, USA},
volume = {16},
number = {1},
issn = {2158-656X},
url = {https://doi.org/10.1145/3706119},
doi = {10.1145/3706119},
journal = {ACM Trans. Manage. Inf. Syst.},
month = feb,
articleno = {6},
numpages = {30}
}

@article{wang2024finlora,
  title={FinLoRA: Finetuning Quantized Financial Large Language Models Using Low-Rank Adaptation},
  author={Wang, Dannong and Kim, Daniel and Jin, Bo and Zhao, Xingjian and Fu, Tianfan and Yang, Steve and Liu, Xiao-Yang},
  journal={arXiv preprint arXiv:2412.11378},
  year={2024}
}

@article{rao2025llms,
  title={LLMs Meet Finance: Fine-Tuning Foundation Models for the Open FinLLM Leaderboard},
  author={Rao, Varun and Sun, Youran and Kumar, Mahendra and Mutneja, Tejas and Mukherjee, Agastya and Yang, Haizhao},
  journal={arXiv preprint arXiv:2504.13125},
  year={2025}
}

@misc{wang2025finlora,
      title={FinLoRA: Benchmarking LoRA Methods for Fine-Tuning LLMs on Financial Datasets}, 
      author={Dannong Wang and Jaisal Patel and Daochen Zha and Steve Y. Yang and Xiao-Yang Liu},
      year={2025},
      eprint={2505.19819},
      archivePrefix={arXiv},
      primaryClass={cs.CE},
      url={https://arxiv.org/abs/2505.19819}, 
}

@inproceedings{lobo2025impact,
  title={On the impact of fine-tuning on chain-of-thought reasoning},
  author={Lobo, Elita and Agarwal, Chirag and Lakkaraju, Himabindu},
  booktitle={Proceedings of the 2025 Conference of the Nations of the Americas Chapter of the Association for Computational Linguistics: Human Language Technologies (Volume 1: Long Papers)},
  pages={11679--11698},
  year={2025}
}

@inproceedings{ye2024fedllm,
author = {Ye, Rui and Ge, Rui and Zhu, Xinyu and Chai, Jingyi and Du, Yaxin and Liu, Yang and Wang, Yanfeng and Chen, Siheng},
title = {FedLLM-bench: realistic benchmarks for federated learning of large language models},
year = {2024},
isbn = {9798331314385},
publisher = {Curran Associates Inc.},
address = {Red Hook, NY, USA},
booktitle = {Proceedings of the 38th International Conference on Neural Information Processing Systems},
articleno = {3528},
numpages = {25},
location = {Vancouver, BC, Canada},
series = {NIPS '24}
}

@article{gao2025flowertune,
  title={Flowertune: A cross-domain benchmark for federated fine-tuning of large language models},
  author={Gao, Yan and Scamarcia, Massimo Roberto and Fernandez-Marques, Javier and Naseri, Mohammad and Ng, Chong Shen and Stripelis, Dimitris and Li, Zexi and Shen, Tao and Bai, Jiamu and Chen, Daoyuan and others},
  journal={arXiv preprint arXiv:2506.02961},
  year={2025}
}

@article{yan2025federated,
  author={Yan, Na and Su, Yang and Deng, Yansha and Schober, Robert},
  journal={IEEE Communications Magazine}, 
  title={Federated Fine-Tuning of LLMs: Framework Comparison and Research Directions}, 
  year={2025},
  volume={63},
  number={10},
  pages={52-58},
  doi={10.1109/MCOM.001.2400770}}

@article{chen2025flow,
  title={Flow of Knowledge: Federated Fine-Tuning of LLMs in Healthcare under Non-IID Conditions},
  author={Chen, Zeyu and Ji, Yun and Wang, Bowen and Shi, Liwen and Zeng, Zijie and Zhang, Sheng},
  journal={arXiv preprint arXiv:2510.00543},
  year={2025}
}

@inproceedings{hilmkil2021scaling,
  title={Scaling federated learning for fine-tuning of large language models},
  author={Hilmkil, Agrin and Callh, Sebastian and Barbieri, Matteo and S{\"u}tfeld, Leon Ren{\'e} and Zec, Edvin Listo and Mogren, Olof},
  booktitle={International Conference on Applications of Natural Language to Information Systems},
  pages={15--23},
  year={2021},
  organization={Springer}
}

@article{hu2022lora,
  title={LoRA: Low-rank adaptation of large language models.},
  author={Hu, Edward J and Shen, Yelong and Wallis, Phillip and Allen-Zhu, Zeyuan and Li, Yuanzhi and Wang, Shean and Wang, Liang and Chen, Weizhu and others},
  journal={Iclr},
  volume={1},
  number={2},
  pages={3},
  year={2022}
}

@article{dettmers2023qlora,
  title={QLoRA: Efficient Finetuning of Quantized LLMs},
  author={Dettmers, Tim and Pagnoni, Artidoro and Holtzman, Ari and Zettlemoyer, Luke},
  journal={Advances in neural information processing systems},
  volume={36},
  pages={10088--10115},
  year={2023}
}

@article{liu2024large,
  title={Large language models in the clinic: a comprehensive benchmark},
  author={Liu, Fenglin and Li, Zheng and Zhou, Hongjian and Yin, Qingyu and Yang, Jingfeng and Tang, Xianfeng and Luo, Chen and Zeng, Ming and Jiang, Haoming and Gao, Yifan and others},
  journal={arXiv preprint arXiv:2405.00716},
  year={2024}
}

@article{ouyang2022training,
  title={Training language models to follow instructions with human feedback},
  author={Ouyang, Long and Wu, Jeffrey and Jiang, Xu and Almeida, Diogo and Wainwright, Carroll and Mishkin, Pamela and Zhang, Chong and Agarwal, Sandhini and Slama, Katarina and Ray, Alex and others},
  journal={Advances in neural information processing systems},
  volume={35},
  pages={27730--27744},
  year={2022}
}

@misc{lin2025open,
      title={Open FinLLM Leaderboard: Towards Financial AI Readiness}, 
      author={Shengyuan Colin Lin and Felix Tian and Keyi Wang and Xingjian Zhao and Jimin Huang and Qianqian Xie and Luca Borella and Matt White and Christina Dan Wang and Kairong Xiao and Xiao-Yang Liu Yanglet and Li Deng},
      year={2025},
      eprint={2501.10963},
      archivePrefix={arXiv},
      primaryClass={cs.CE},
      url={https://arxiv.org/abs/2501.10963}, 
}

@inproceedings{liu2022few,
author = {Liu, Haokun and Tam, Derek and Muqeeth, Mohammed and Mohta, Jay and Huang, Tenghao and Bansal, Mohit and Raffel, Colin},
title = {Few-shot parameter-efficient fine-tuning is better and cheaper than in-context learning},
year = {2022},
isbn = {9781713871088},
publisher = {Curran Associates Inc.},
address = {Red Hook, NY, USA},
booktitle = {Proceedings of the 36th International Conference on Neural Information Processing Systems},
articleno = {142},
numpages = {16},
location = {New Orleans, LA, USA},
series = {NIPS '22}
}

@article{han2024parameter,
  title={Parameter-efficient fine-tuning for large models: A comprehensive survey},
  author={Han, Zeyu and Gao, Chao and Liu, Jinyang and Zhang, Jeff and Zhang, Sai Qian},
  journal={arXiv preprint arXiv:2403.14608},
  year={2024}
}

@misc{2025II-Medical-8B-1706,
      title={II-Medical-8B: Medical Reasoning Model}, 
      author={Intelligent Internet},
      year={2025}
}

@misc{qwen3technicalreport,
      title={Qwen3 Technical Report}, 
      author={Qwen Team},
      year={2025},
      eprint={2505.09388},
      archivePrefix={arXiv},
      primaryClass={cs.CL},
      url={https://arxiv.org/abs/2505.09388}, 
}

@misc{qwen2.5,
    title = {Qwen2.5: A Party of Foundation Models},
    url = {https://qwenlm.github.io/blog/qwen2.5/},
    author = {Qwen Team},
    month = {September},
    year = {2024}
}

@misc{llama3modelcard,
    title={Llama 3 Model Card},
    author={AI@Meta},
    year={2024},
    url = {https://github.com/meta-llama/llama3/blob/main/MODEL_CARD.md}
}

@misc{yu2025finemedlm,
    title={FineMedLM-o1: Enhancing the Medical Reasoning Ability of LLM from Supervised Fine-Tuning to Test-Time Training}, 
    author={Hongzhou Yu and Tianhao Cheng and Ying Cheng and Rui Feng},
    year={2025},
    eprint={2501.09213},
    archivePrefix={arXiv},
    primaryClass={cs.CL},
    url={https://arxiv.org/abs/2501.09213}, 
}

@misc{gemma_2024,
    title={Gemma},
    url={https://www.kaggle.com/m/3301},
    DOI={10.34740/KAGGLE/M/3301},
    publisher={Kaggle},
    author={Gemma Team},
    year={2024}
}

@misc{gemma_2025,
    title={Gemma 3},
    url={https://goo.gle/Gemma3Report},
    publisher={Kaggle},
    author={Gemma Team},
    year={2025}
}

@article{grattafiori2024llama3herd,
  title   = {The Llama 3 Herd of Models},
  author  = {Grattafiori, Aaron and Dubey, Abhimanyu and Jauhri, Abhinav and Pandey, Abhinav and Kadian, Abhishek and Al-Dahle, Ammar and Letman, Adam and Mathur, Anant and Schelten, Alan and Vaughan, Angela and others},
  journal = {arXiv preprint arXiv:2407.21783},
  year    = {2024}
}

@misc{medicalllmleaderboard2024,
  title        = {The Open Medical-LLM Leaderboard: Benchmarking Large Language Models in Healthcare},
  author       = {Ura, Aaditya and Minervini, Pasquale and Fourrier, Cl{\'e}mentine},
  year         = {2024},
  month        = apr,
  howpublished = {\url{https://huggingface.co/blog/leaderboard-medicalllm}},
  note         = {Hugging Face blog post, accessed 2026-04-10}
}

@article{schwartz2020green,
  title={Green AI},
  author={Schwartz, Roy and Dodge, Jesse and Smith, Noah A and Etzioni, Oren},
  journal={Communications of the ACM},
  volume={63},
  number={12},
  pages={54--63},
  year={2020},
  publisher={ACM New York, NY, USA}
}

\end{document}